\pdfoutput=1
\documentclass[10pt, conference, compsocconf]{IEEEtran}
\ifCLASSINFOpdf
\else
\fi

\usepackage{cite}
\usepackage{hyperref}
\usepackage{amsthm,amsmath,amssymb,amsfonts}
\usepackage{array}
\usepackage{graphicx}
\usepackage{textcomp}
\usepackage{xcolor}
\usepackage[ruled]{algorithm2e}
\usepackage{algpseudocode}
\usepackage{subfigure}
\usepackage{multirow}
\usepackage{pifont}
\usepackage{booktabs}
\usepackage{float}
\definecolor{mypurple}{RGB}{153,51,255}
\definecolor{myblue}{RGB}{0,102,255}

\newtheorem{myDef}{Definition}

\begin{document}
%
\title{ACE-HGNN: Adaptive Curvature Exploration Hyperbolic Graph Neural Network}

\author{\IEEEauthorblockN{Xingcheng Fu\IEEEauthorrefmark{1}\IEEEauthorrefmark{2},
Jianxin Li\IEEEauthorrefmark{1}\IEEEauthorrefmark{2},
Jia Wu\IEEEauthorrefmark{3}, 
Qingyun Sun\IEEEauthorrefmark{1}\IEEEauthorrefmark{2}, 
Cheng Ji\IEEEauthorrefmark{1}\IEEEauthorrefmark{2},
Senzhang Wang\IEEEauthorrefmark{4},
Jiajun Tan\IEEEauthorrefmark{2},  
Hao Peng\IEEEauthorrefmark{1} and
Philip S. Yu\IEEEauthorrefmark{5}}
\IEEEauthorblockA{\IEEEauthorrefmark{1}Beijing Advanced Innovation Center for Big Data and Brain Computing, Beihang University, Beijing 100191, China}
\IEEEauthorblockA{\IEEEauthorrefmark{2}School of Computer Science and Engineering, Beihang University, Beijing 100191, China}
\IEEEauthorblockA{\IEEEauthorrefmark{3}Department of Computing, Macquarie University, Sydney, Australia}
\IEEEauthorblockA{\IEEEauthorrefmark{4}School of Computer Science and Engineering, Central South University, Changsha 410083, China}
\IEEEauthorblockA{\IEEEauthorrefmark{5}Department of Computer Science, University of Illinois at Chicago, Chicago, IL 60607, USA}
 Email: \{fuxc, lijx, sunqy, jicheng, penghao\}@act.buaa.edu.cn, jia.wu@mq.edu.au,\\ szwang@csu.edu.cn, chiachiun\_than@buaa.edu.cn,psyu@uic.edu}

\maketitle

\begin{abstract}
Graph Neural Networks (GNNs) have been widely studied in various graph data mining tasks. 
Most existing GNNs embed graph data into Euclidean space and thus are less effective to capture the ubiquitous hierarchical structures in real-world networks. 
Hyperbolic Graph Neural Networks (HGNNs) extend GNNs to hyperbolic space and thus are more effective to capture the hierarchical structures of graphs in node representation learning. 
In hyperbolic geometry, the graph hierarchical structure can be reflected by the curvatures of the hyperbolic space, and different curvatures can model different hierarchical structures of a graph. 
However, most existing HGNNs manually set the curvature to a fixed value for simplicity, which achieves a suboptimal performance of graph learning due to the complex and diverse hierarchical structures of the graphs. 
To resolve this problem, we propose an Adaptive Curvature Exploration Hyperbolic Graph Neural Network named ACE-HGNN to adaptively learn the optimal curvature according to the input graph and downstream tasks. 
Specifically, ACE-HGNN exploits a multi-agent reinforcement learning framework and contains two agents, ACE-Agent and HGNN-Agent for learning the curvature and node representations, respectively. The two agents are updated by a Nash Q-leaning algorithm collaboratively, seeking the optimal hyperbolic space indexed by the curvature. 
Extensive experiments on multiple real-world graph datasets demonstrate a significant and consistent performance improvement in model quality with competitive performance and good generalization ability. 

\end{abstract}


\IEEEpeerreviewmaketitle

\section{Introduction}
Graphs are widely used to model the complex relationships between objects, such as social networks~\cite{hamilton2017inductive}, academic networks~\cite{maulik2006advanced}, and biological networks~\cite{gilmer2017neural}. 
In recent years, graph representation learning has shown its effectiveness in capturing the irregular but related complex structures in graph data~\cite{hamilton2017inductive}. 
As an important topology characteristic, hierarchical structures are ubiquitous in many real-world graphs, such as the hypernym structure in natural languages~\cite{NickelK17Poincare,PoincareGlove}, the subordinate structure of entities in the knowledge graph~\cite{balazevic2019multi,wang2020h2kgat}, and the organizational structure for fraud detection~\cite{dou2020enhancing} or complex production systems~\cite{ren2020data,ren2020wide}. 
In our work, we focus on studying the tree-like structure\footnote{We interchangeably use the terms \textit{hierarchy} or \textit{hierarchical structure} and \textit{tree-like structure} in this paper. } which exists extensively in real-world networks~\cite{Krioukov2010Hyperbolic}.

\begin{figure}[t]
	\centering
	\includegraphics[width=0.48\textwidth]{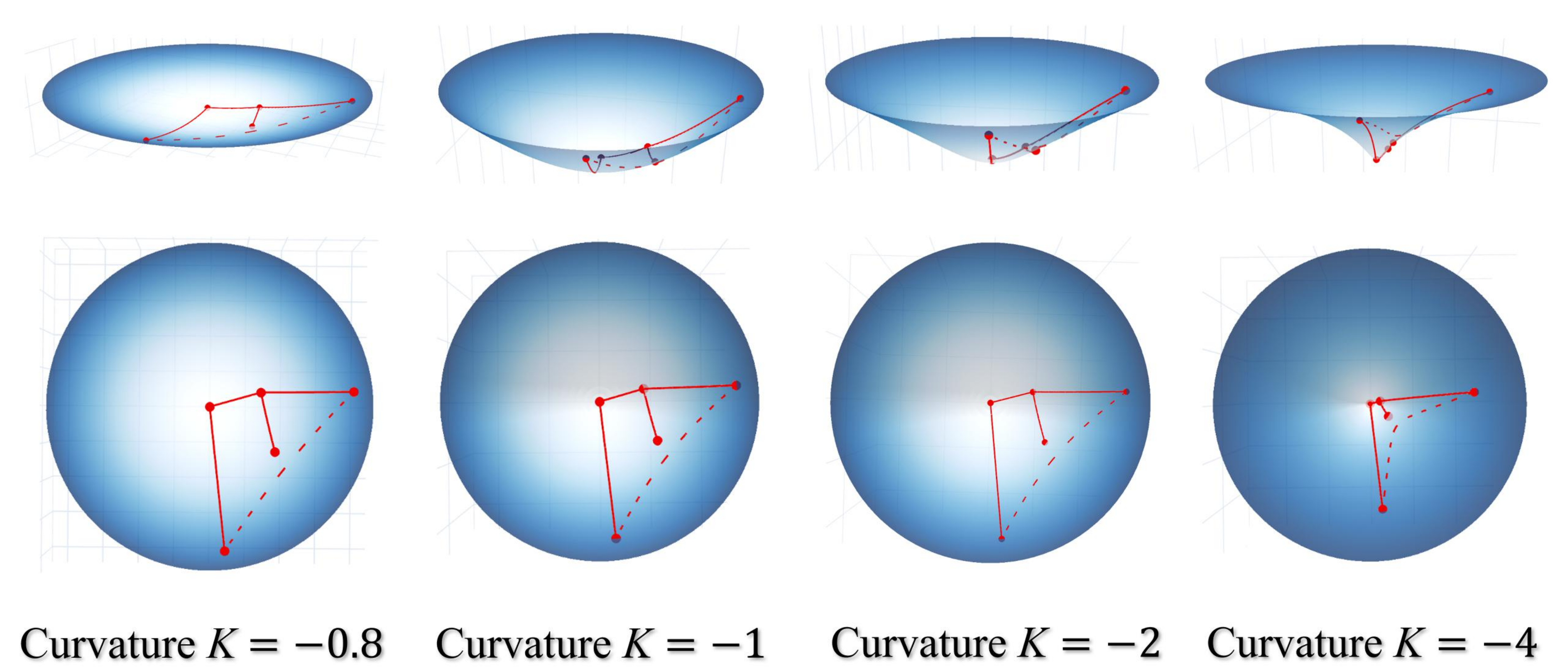}
    \vspace{-1em}
	\caption{An example of hyperbolic graph distance (\textcolor{red}{red solid line}) and hyperbolic embedded distance (\textcolor{red}{red dash line}) between two nodes on a binary tree in hyperbolic spaces with different curvatures ($K$). 
}
\vspace{-2em}
\label{example_1}
\end{figure}

Recently, hyperbolic space has been introduced into Graph Neural Networks (GNNs) and achieves better performance and interpretability in node representation learning, because compared with the Euclidean space, the hierarchical structure of graphs can be better captured in the hyperbolic space~\cite{NickelK17Poincare}. 
To learn the node representations in hyperbolic space, Hyperbolic Graph Neural Networks (HGNNs) extend the node feature aggregation of Graph Neural Networks into hyperbolic space, and effectively fuse the node features and the hierarchical structures to acquire the superior node representations for graphs. 
However, even though most real-world graphs have tree-like structures, how a graph is like a tree differs significantly. 
Thus, a major limitation of existing HGNNs is that they cannot adaptively choose the appropriate hyperbolic space to represent the graphs with different degrees of tree-like structures.

Inspired by some relevant works~\cite{cannon1997hyperbolic,Krioukov2010Hyperbolic}, we use hyperbolic curvature to measure similarity between hyperbolic geometry and Euclidean geometry.
In addition, some recent works~\cite{gu2019learning,bachmann2020constant} on graph representation learning have focused on the relationship between graph structures and geometric embedding spaces with different curvatures. 
Graphs with different structures (e.g., cycle or tree structure) have different information distortion when embedded in the Euclidean or non-Euclidean embedding spaces. 
\figurename~\ref{example_1} illustrates an intrinsic connection between hyperbolic curvatures and hierarchies of graphs. 
It shows that different curvatures significantly affect distance metrics in hyperbolic space. 
When the curvature decreases, the hyperbolic embedded distance is more reflective of the tree structure because it is close to the shortest path length of the two nodes (i.e., hyperbolic graph distance). 
When the curvature approaching zero, the hyperbolic embedded distance is close to the Euclidean embedded distance, leading to the information loss of the hierarchical structure. 
Therefore, a natural problem is, “\textit{Can we design a Hyperbolic Graph Neural Network model which can automatically learn the appropriate curvature to obtain desirable adaptability for different graphs with tree-like structures?}” 

In existing hyperbolic graph learning methods, the curvature is either chosen as a hyper-parameter or learned in the training process. 
$\kappa$GCN~\cite{bachmann2020constant} and mixed product spaces learning~\cite{gu2019learning} try to adjust curvature only depending on the estimation of the original graph topology, which cannot adapt to different graphs and downstream tasks. 
HGCN~\cite{HGCN_ChamiYRL19} takes curvature as a parameter to learn. 
However, they only slightly adjust the curvature in a relatively limited range to make the node representation learning process converge. 

In conclusion, there still lacks an unified and end-to-end learning architecture for both graph representation learning and optimal curvature searching. 

To address the above problems, we propose a novel \underline{\textbf{A}}daptive \underline{\textbf{C}}urvature \underline{\textbf{E}}xploration \underline{\textbf{H}}yperbolic \underline{\textbf{G}}raph \underline{\textbf{N}}eural \underline{\textbf{N}}etwork named \textbf{ACE-HGNN}\footnote{The datasets and the source code are released at \url{https://github.com/RingBDStack/ACE-HGNN}. }. 
The basic idea is that we take the downstream task as the environment in reinforcement learning (RL) and calculate the feedback reward based on the hyperbolic node representations. 
Specifically, we design two agents: the adaptive curvature exploration agent (\textbf{ACE-Agent}) and the hyperbolic graph neural network agent (\textbf{HGNN-Agent}). 
The ACE-Agent is designed to independently learn the optimal curvature in a broad range of the parameter space based on reinforcement learning. 
The HGNN-Agent with variable curvature is designed to learn the node representations in hyperbolic space with particular curvature. 
To fully ensure learning both curvature exploration and graph representations in hyperbolic space, we propose a multi-agent reinforcement learning framework to search for the optimal curvature and learn node representations collaboratively. 
The optimization objective is that the two cooperative agents achieve a Nash equilibrium. 
The visualization results in hyperbolic space provide an intuitive understanding of how different curvatures affect the model's capability of capturing the hierarchy. 
We summarize our contributions as follows: 
\begin{itemize}
\item We make an observation and analysis of the adaptability problem for HGNNs with different hierarchical topologies in hyperbolic space, and transform the embedding space adaptability problem into an optimal curvature exploration problem in hyperbolic space. 
\item We propose a novel end-to-end architecture, named adaptive curvature exploration hyperbolic graph neural network (ACE-HGNN), to guide the selection of optimal embedding space. 
To our best knowledge, it is the first attempt to utilize reinforcement learning in hyperbolic machine learning. 
\item Extensive experiments on five typical real-world datasets demonstrate a significant and consistent improvement in model adaptability with competitive performance. And we also visualize the node representations and attention scores of ACE-HGNN to intuitively show the capability of capturing structures.

\end{itemize}

\section{Preliminaries and Notations}
\label{sec:background}
Hyperbolic space commonly refers to manifolds with constant negative curvature and is used for modeling complex networks~\cite{Krioukov2010Hyperbolic}. 
Among the common isometric models used to describe hyperbolic spaces~\cite{cannon1997hyperbolic}, the hyperboloid (Lorenz) model has recently been widely used in machine learning~\cite{HNN:GaneaBH18,HGNN_Qi,HGCN_ChamiYRL19}.  


\textbf{Hyperboloid manifold.}
The hyperboloid model is an $n$-dimensional hyperbolic geometry as a manifold in the $(n\!+\!1)$-dimensional Minkowski space. 
A hyperboloid manifold $\mathbb{H}^{n,\zeta}\!=\!\{ \mathbf{x}\!\in\!\mathbb{R}^{n+1}\!\mid\!\langle \mathbf{x},\mathbf{x} \rangle_{\mathbb{H}}\!=\!-1/\zeta^2, \zeta\!>\!0 \}$ in $n$-dimensional space with curvature\footnote{In this paper, we consider hyperbolic space is a constant negative curvature space, and the curvature is the sectional curvature~\cite{gu2019learning}. 
To facilitate the calculation and expression, we interchangeably use the curvature $K$ and the curvature parameter $\zeta$. 
} $K\!=\!-1/\zeta^2$ ($\zeta> 0$). 
$\langle \cdot,\cdot \rangle_\mathbb{H}$ is Lorentzian scalar product and $\langle \mathbf{x},\mathbf{y} \rangle_\mathbb{H} := -x_0 y_0 + x_1 y_1 + \dots + x_n y_n$. 
The distance $\mathrm{d}^{\zeta}_{\mathbb{H}}(\mathbf{x},\mathbf{y})$ on the hyperboloid model is defined as: 
\begin{equation}
\label{equ:2}
\begin{aligned}
     d^{\zeta}_{\mathbb{H}}\left (\mathbf{x},\mathbf{y}\right ) = \zeta\mathrm{arccosh}\left (\langle \mathbf{x},\mathbf{y} \rangle_\mathbb{H} / \zeta^2\right ). 
\end{aligned}
\end{equation}


\textbf{Logarithmic map and exponential map.}
For $\mathbf{x}, \mathbf{y} \in \mathbb{H}^{n,\zeta}$, $\mathbf{v} \in \mathcal{T}_{\mathbf{x}}\mathbb{H}^{n,\zeta}$, such $\mathbf{v} \neq 0$ and $\mathbf{x} \neq \mathbf{y}$, $\mathbf{x}$ is the reference point, the logarithmic map is a map from hyperboloid space $\mathbb{H}^{n,\zeta}$ to the tangent space $\mathcal{T}_{\mathbf{x}}\mathbb{H}^{n,\zeta}$, and the exponential map maps back to hyperboloid space $\mathbb{H}^{n,\zeta}$. 
The tangent space $\mathcal{T}_{\mathbf{x}}\mathbb{H}^{n,\zeta}:=\{\mathbf{v} \in \mathbb{R}^{n+1}|\langle \mathbf{v}, \mathbf{x} \rangle_{\mathbb{H}}=0\}$ centered at point $\mathbf{x}$ is a $n$-dimensional Euclidean space. 
The logarithmic map $\mathrm{log}^{\zeta}_{\mathbf{x}}(\cdot)$ and exponential map $\mathrm{exp}^{\zeta}_{\mathbf{x}}(\cdot)$ are:
\begin{align}
\mathrm{log}^{\zeta}_{\mathbf{x}}(\mathbf{y}) &= d^{\zeta}_{\mathbb{H}}\left (\mathbf{x},\mathbf{y}\right ) \frac{\mathbf{y}+\frac{1}{\zeta^2}\langle\mathbf{x},\mathbf{y}\rangle_\mathbb{H} \mathbf{x}}{\| \mathbf{y}+\frac{1}{\zeta^2}\langle\mathbf{x},\mathbf{y}\rangle_\mathbb{H} \mathbf{x} \|_{\mathbb{H}}},\label{eq:log}\\
     \mathrm{exp}^{\zeta}_{\mathbf{x}}(\mathbf{v}) &= \mathrm{cosh} \left (\frac{ \| \mathbf{v} \|_{\mathbb{H} } }{\zeta }\right)\mathbf{x} + \zeta \mathrm{sinh} \left (\frac{ \| \mathbf{v} \|_{\mathbb{H} } }{\zeta }\right ) \frac{\mathbf{v}}{\|\mathbf{v}\|_{\mathbb{H}}} , \label{eq:exp}
\end{align}
where $\| \mathbf{v} \|_{\mathbb{H}} = \sqrt{\langle \mathbf{v}, \mathbf{v} \rangle_{\mathbb{H}}}$ is the Lorentzian norm of $\mathbf{v}$.


\begin{figure*}[t]
\centering
\subfigure[Distances in Poincaré disk.]{
\includegraphics[width=0.21\linewidth]{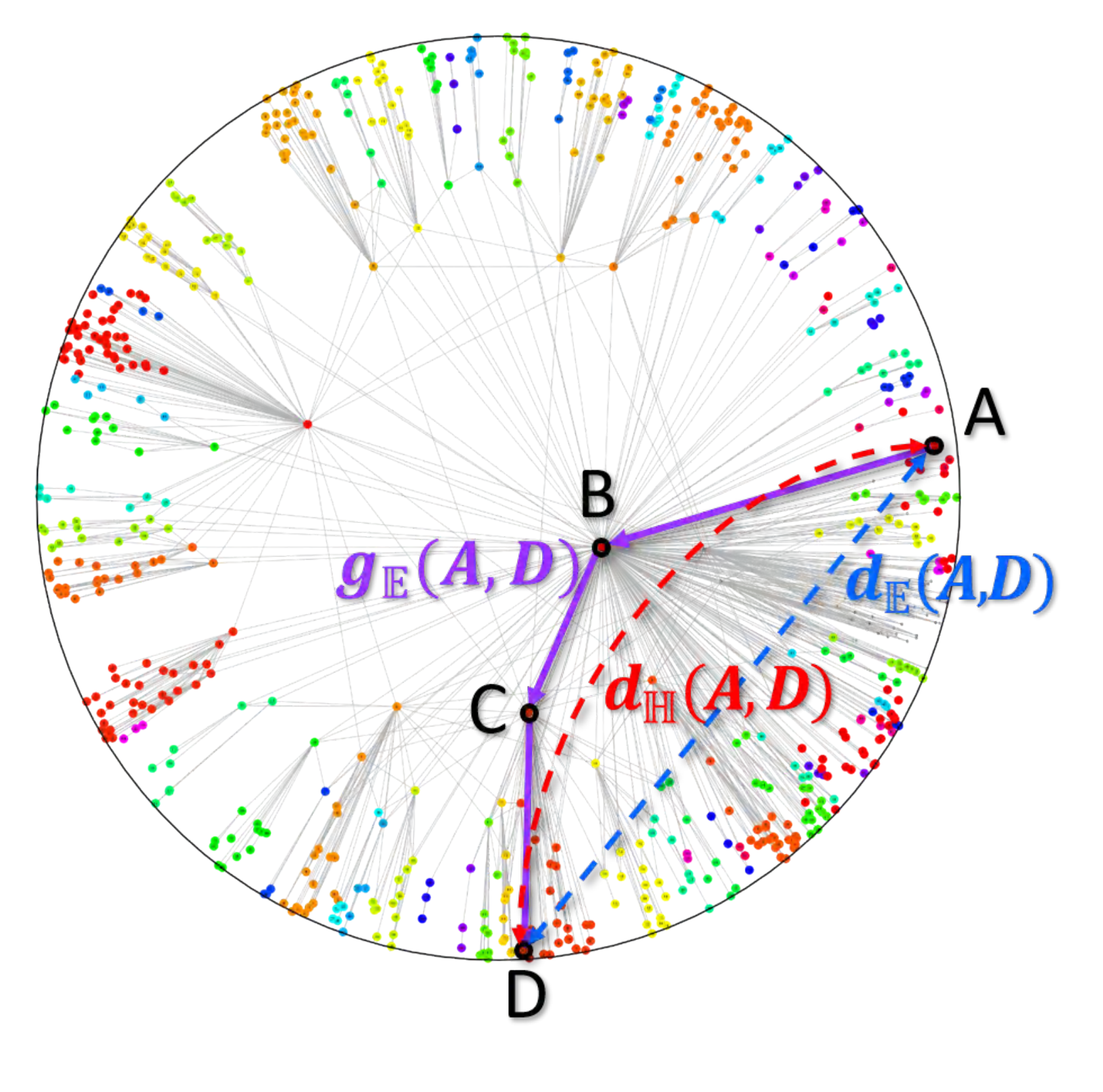}
\label{example:a}
}%
\hspace{15pt}
\subfigure[Distances projection.]{
\includegraphics[width=0.22\linewidth]{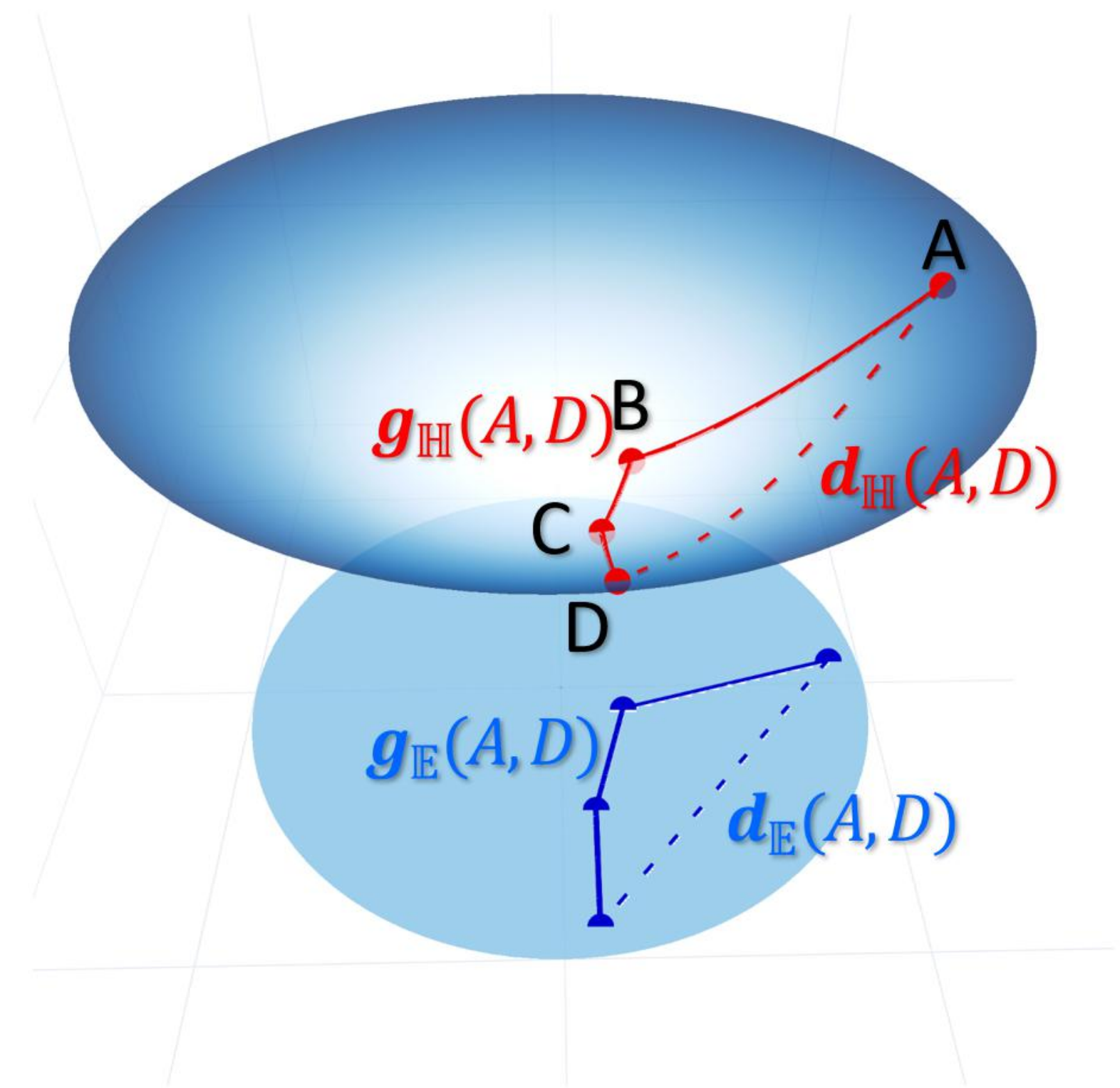}
\label{example:b}
}%
\hspace{15pt}
\subfigure[Distances with different curvature.]{
\includegraphics[width=0.45\linewidth]{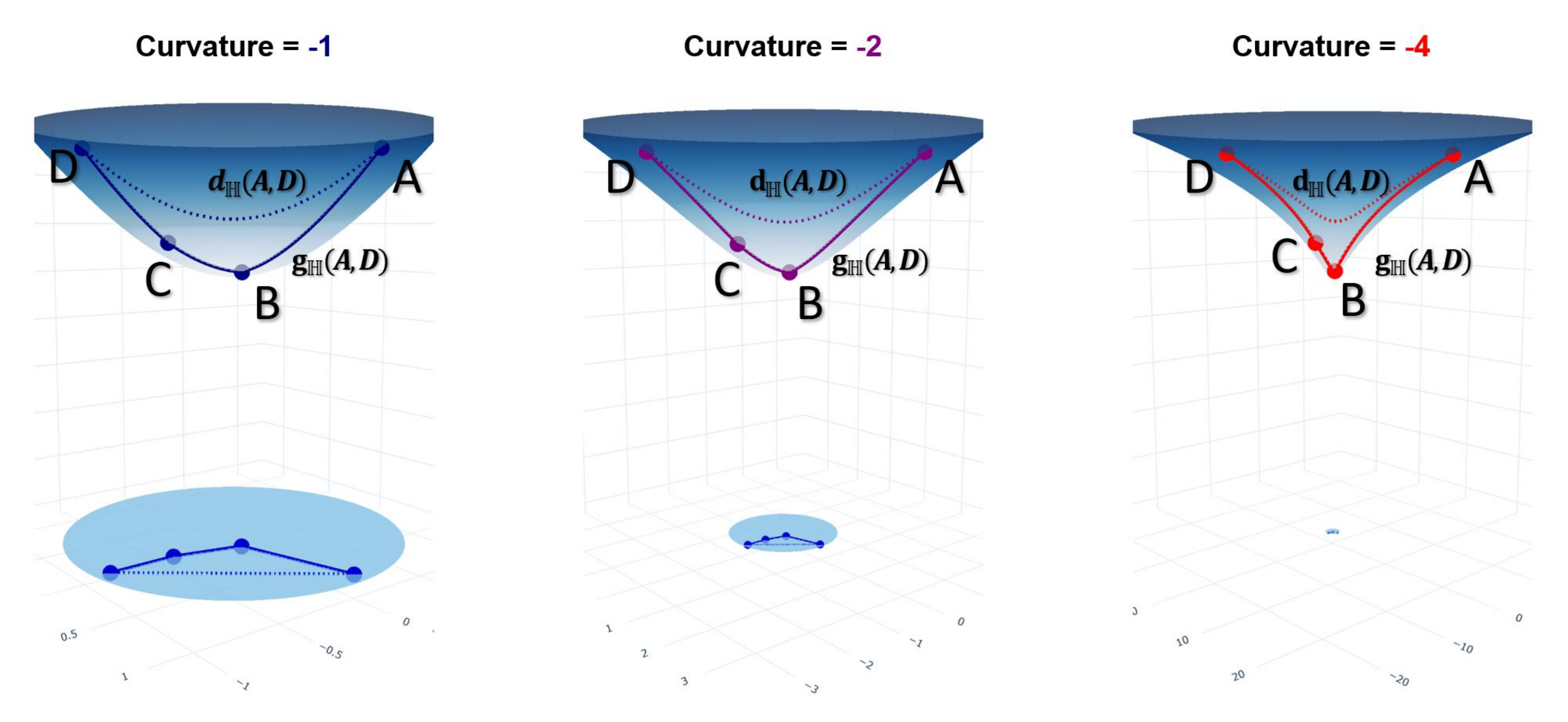}
\label{example:c}
}
\centering
\vspace{-0.5em}
\caption{An illustration of different distance metrics in hyperbolic spaces with different curvature. 
(a) Graph distance (\textcolor{mypurple}{purple solid lines}), Euclidean distance (\textcolor{myblue}{blue dashed line}) and hyperbolic geodesics (\textcolor{red}{red dashed curve}) on a tree-like graph in Poincaré disk. 
(b) Graph distance and embedded distance in Poincaré disk (Euclidean projection, \textcolor{myblue}{blue solid and dashed lines}) and hyperboloid (curvature $K\!=\!-1$, \textcolor{red}{red solid and dashed curves}). 
(c) Graph distance and hyperbolic distance on the hyperboloid of different curvature. }
\label{example}
\vspace{-1em}
\end{figure*}

\section{Curvature versus Hierarchical Structure}
\label{sec:curv}
In this section, we perform a qualitative analysis and leverage the embedding distortion to reveal the intrinsic connections between curvature and the hierarchical structure. 
Then, we give a further analysis to demonstrate curvature is the dominant factor of model's representation ability. 
Finally, we transform the adaptability problem of hyperbolic graph representation learning into a multi-objective optimization problem of curvature exploration and model optimization. 

\subsection{Embedding Distortion}
Previous works use distortion to measure the information loss of topology embedding in Euclidean space~\cite{gu2019learning,bachmann2020constant}. 
However, the feature information, which plays an important role in graph representation learning, is rarely taken into account by the current distortion measurements. 
Therefore, to measure the distortion of both hierarchical structure and feature information, we propose the embedding distortion to measure the embedding quality in hyperbolic space. 
First, we define two important distances including embedded distance and graph distance as follows. 

\textbf{Embedded distance} is the distance between two nodes in the embedding space (dash lines in \figurename~\ref{example}), which can be regarded as the semantic distance or feature similarity between two nodes. 
We use $d_{\mathbb{E}}$ and $d_{\mathbb{H}}$ to denote the embedding distance in Euclidean and hyperbolic embedding spaces, which can be computed by the inner product and the Lorentzian scalar product, respectively. 
For the nodes $A$ and $D$ in \figurename~\ref{example:a}, their Euclidean embedded distance is $d_{\mathbb{E}}(A,D)$ (\textcolor{myblue}{blue dash line}), and the hyperbolic embedded distance is $d_{\mathbb{H}}(A,D)$ (\textcolor{red}{red dash curve}). 

\textbf{Graph distance} is the length of the shortest path between two nodes in a graph. 
Information propagation is an essential function of many real-world networks (e.g., Internet and brain network), and we use it to measure the topological distance between two nodes in graph representation learning. 
As shown in \figurename~\ref{example:a}, the hyperbolic graph distance between nodes $A$ and $D$ (\textcolor{mypurple}{purple solid lines}) is $g_{\mathbb{H}}(A,D)=d_{\mathbb{H}}(A,B)+d_{\mathbb{H}}(B,C)+d_{\mathbb{H}}(C,D)$. 

Embedding distortion can be defined as the overall inconsistency between embedded distance and graph distance.
Formally, we define embedding distortion as follows. 
\begin{myDef}[\textbf{Embedding Distortion}]\label{def:distortion}
Given a graph $G$ with node set $V$, 
the embedding distortion $\mathcal{D}$ in the hyperbolic embedding space $\mathbb{H}$ is defined as: 
\begin{equation}
\label{eq:distortion}
   \mathcal{D} = \frac{1}{|V^2|} \sum_{i,j \in V} \left | \left(\frac{d_{\mathbb{H}}^2(i,j)}{g_{\mathbb{H}}^2(i,j)}-1 \right) \right |.
\end{equation}
\end{myDef}

\subsection{Connection of Curvature and Embedding Distortion}
In this section, we analyze the connection between embedding distortion and curvature by showing how hyperbolic distance changes with curvature. 
We employ the following conclusions from the existing work of hyperbolic geometry in complex network~\cite{Krioukov2010Hyperbolic} to analyze the hyperbolic distance metrics. 
The hyperbolic distance (in hyperboloid manifold) $d_{\mathbb{H}}$ between two points at polar coordinates $(r, \theta)$ and $(r^{\prime}, \theta^{\prime})$ is given by the hyperbolic law of cosines:
\begin{equation}\label{hyperdist1}
   \cosh \frac{d_{\mathbb{H}}}{\zeta} =\cosh \frac{r}{\zeta}  \cosh \frac{r^{\prime}}{\zeta} -\sinh \frac{r}{\zeta} \sinh \frac{r^{\prime}}{\zeta} \cos \Delta \theta,
\end{equation}
where $\Delta\theta$ is the angle between the points.  Eq.~\eqref{hyperdist1} converges to their familiar Euclidean analogs ($d_{\mathbb{H}} \!\rightarrow\! d_{\mathbb{E}}$ at $\zeta \!\rightarrow\! \infty$). 
For sufficiently large $r/\zeta, r^{\prime}/\zeta$, and $\Delta\theta \!>\! 2\sqrt{e^{-2r/\zeta}\!-\!e^{-2r^{\prime}/\zeta}}$, $d_{\mathbb{H}}$ can be closely approximated by: 
\begin{equation}\label{hyperdist2}
   d_{\mathbb{H}}=r+r^{\prime}+2\zeta \ln \sin \frac{\Delta \theta}{2} \approx r+r^{\prime}+2\zeta \ln \frac{\Delta \theta}{2}.
\end{equation}
The hyperbolic embedded distance $d_{\mathbb{H}}$ between two points is approximately the sum of their radius, minus some $\Delta\theta$-dependent correction, and $\Delta\theta \rightarrow 0$ at $\zeta \rightarrow 0$. 
Then we denote $(r_1, \theta_1),(r_2, \theta_2),\cdots,(r_n, \theta_n)$ as points in the shortest path between $(r, \theta)$ and $(r^{\prime}, \theta^{\prime})$. 
According to Eq.~\eqref{hyperdist2}, we can derive the hyperbolic graph distance $g_{\mathbb{H}}$ as follows:
\begin{equation}\label{hypergraphdist}
   g_{\mathbb{H}}= r + r^{\prime} + 2\sum_{m=1}^{n}r_m + 2\zeta \sum_{k=1}^{n+1}\ln \frac{\Delta \theta_k}{2},
\end{equation}
where $\Delta \theta_k$ is the angle of each adjacent point pair in the shortest path. 
For the tree-like graph, the shortest path navigation between two nodes tends to be close to the center, and this property is enhanced with the curvature parameter $\zeta$ decreasing. 
Based on the above properties of tree-like graph in hyperbolic space,  we can observe $d_{\mathbb{H}} \rightarrow g_{\mathbb{H}}$ when $\zeta \!\rightarrow\! 0$ and $\Delta \theta_k \!\rightarrow\! 0$. 
To give a more intuitive explanation, \figurename~\ref{example} shows a simple structure in the hyperboloid manifold with different curvatures.

Based on the above properties of curvature, we can transform the adaptability problem of hyperbolic graph representation learning as a multi-objective optimization problem consisting of the optimal curvature selection and HGNN optimization, as defined in the next section.


\subsection{Problem Definition }
We aim to minimize the embedding distortion and learn the optimal graph node representation by the HGNNs simultaneously. 
Thus, we take the combination of optimal curvature selection and HGNNs optimization as a multi-objective optimization problem.
Given a graph $G$ with label set $Y$, the problem can be defined as follows:
\begin{equation}
\begin{aligned}
&\mathop{\arg\min}_{\phi,K}\vec{f}(G)=\left [Loss\left (\mathbf{h},Y;\phi,K\right ),\mathcal{D}\left (\mathbf{h};K\right )\right ]\\
&\mathbf{s.t.}~\mathbf{h}=\mathrm{HGNN}(G), K\in (-\infty,0\;].
\end{aligned}
\end{equation}
However, in multi-objective optimization, there is usually no unique global optimal solution~\cite{stadler1979survey}. 
Fortunately, current deep reinforcement learning paradigm provides us with inspiration. 
We attempt to introduce multiple agents as interactive decision makers to solve the multi-objective optimization efficiently for the specific downstream task. 
In the next section, we will introduce our method based on multi-agent reinforcement learning. 

\begin{figure*}[t]
\centering
\includegraphics[width=1\textwidth]{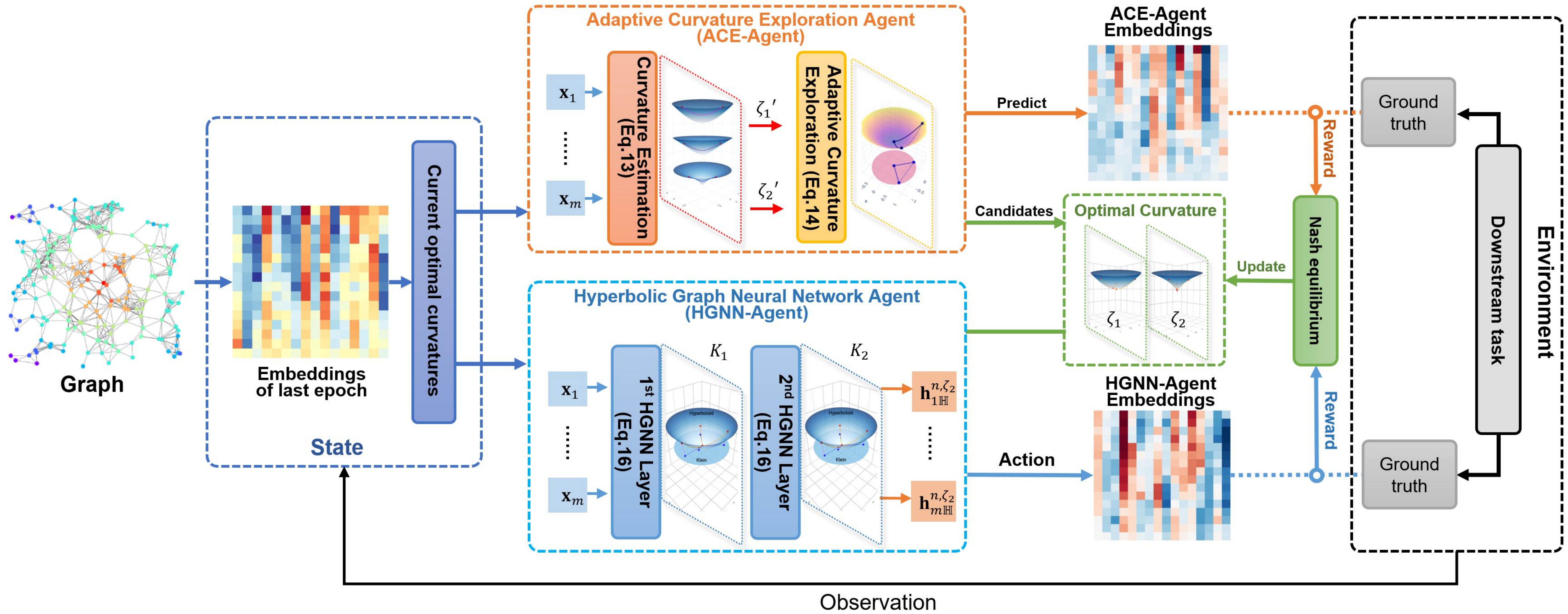}
\vspace{-1em}
\caption{An illustration of ACE-HGNN architecture. 
ACE-HGNN consists of two major components: ACE-Agent and HGNN-Agent. 
The ACE-Agent explores the optimal curvature, and the HGNN-Agent learns representations in hyperbolic space with a given curvature. 
Both the two agents are updated by a multi-agent reinforcement learning algorithm (Nash Q-learning) until they find an embedding space with an appropriate curvature. 
}
\vspace{-1em}
\label{framework}
\end{figure*}

\section{Proposed Methodology}\label{section 4}
In this section, we outline the ACE-HGNN framework, an Adaptive Curvature Exploration Hyperbolic Graph Neural Network, and show how it is applied for hyperbolic node representation learning.
\figurename~\ref{framework} shows our collaborative reinforcement learning framework with two agents to solve the multi-objective optimization problem, where the Adaptive Curvature Exploration Agent (ACE-Agent) explores the curvature for better hyperbolic representation space and Hyperbolic Graph Neural Network Agent (HGNN-Agent) learns the node representations in the hyperbolic space with the selected curvature.

\subsection{Adaptive Curvature Exploration Agent (ACE-Agent)}
Owing to the properties of hyperbolic logarithmic mapping and Riemannian optimization~\cite{DBLP:conf/icml/NickelK18}, the range of curvature updating is fairly small during the learning process (our experiments confirmed this phenomenon in Section~\ref{section 5}). 
Therefore, it is difficult to update it as a learning parameter using back-propagation in the HGNN model. 
To address this issue, we design an Adaptive Curvature Exploration Agent (ACE-Agent), which adaptively explores the optimal curvature under the reinforcement learning framework for each HGNN layer. Formally, we define the state, action, and reward of the ACE-Agent as follows. 

\textbf{ACE-Agent State $S^{t}_{\mathrm{ACE}}$}: 
Since the curvature directly indicates the hyperbolic embedding space, we define the state of ACE-Agent of epoch $t$ as:
\begin{equation}\label{acestate}
\begin{aligned}
    S^{t}_{\mathrm{ACE}} = \left (\zeta^{t-1}_1,\dots,\zeta^{t-1}_L\right ),
\end{aligned}
\end{equation}
where $(\zeta^{t-1}_1,\dots,\zeta^{t-1}_L)$ are the explored curvatures of the last epoch, and $L$ is the number of layers of the model in HGNN-Agent. 
We take the state when the reinforcement learning terminates as the explored optimal curvature. 

\textbf{ACE-Agent Action $A^{t}_{\mathrm{ACE}}$}: 
To minimize the embedding distortion in Eq.~\eqref{eq:distortion} and explore the optimal curvature, we use the deviation from the Parallelogram Law to estimate the graph curvature~\cite{gu2019learning,bachmann2020constant}. 
Let $(a,b,c)$ be a geodesic triangle in hyperboloid space $\mathbb{H}^{n,\zeta}$, and $m$ be the (geodesic) midpoint of $(b,c)$. 
Their quantities are defined as: 
\begin{equation}\label{curvest}
\begin{aligned}
    &\xi_{\mathbb{H}}(a,b;c)\! =\! g_{\mathbb{H}}(a,m)^2\! +\! \frac{g_{\mathbb{H}}(b,c)^2}{4}\! +\! \frac{g_{\mathbb{H}}(a,b)^2+g_{\mathbb{H}}(a,c)^2}{2}, \\
    &\xi_{\mathbb{H}}(m;a,b;c) = \frac{1}{2 g_{\mathbb{H}}(a,m)} \xi_{\mathbb{H}}(a,b;c).
\end{aligned}
\end{equation}
We design our curvature updating according to Eq.~\eqref{curvest}. 
The new average curvature estimation $\kappa^{t}$ in epoch $t$ is defined as:
\begin{equation}
\label{newcurv}
\begin{aligned}
    \kappa^{t} &= \frac{1}{|V|}\sum_{m\in V}\left (\frac{1}{n_s}\sum_{j=0}^{n_s}\xi_{\mathbb{H}}\left (\mathbf{h}^{t-1}_{m};\mathbf{h}^{t-1}_{a_j},\mathbf{h}^{t-1}_{b_j};\mathbf{h}^{t-1}_{c_j}\right )\right ), 
\end{aligned}
\end{equation}
where $b$ and $c$ are randomly sampled from the neighbors of $m$, and $a$ is a node in the graph $G$ except for $\{m, b, c\}$. 
For each node, we perform the above sampling $n_s$ times and take the average value as the new estimated curvature. 

Then we transform the embeddings $\mathbf{h}^{\zeta^{t-1},t-1}$ of epoch $t-1$ from HGNN-Agent as the input into the hyperboloid manifold with the new curvature $\zeta^{t}$: 
\begin{equation}
\label{eq:mapcurv}
\begin{aligned}
    \zeta^{t} &= \left (1-\gamma\right )\zeta^{t-1} + \frac{\gamma}{\sqrt{-\kappa^{t}}}, \\
    \mathbf{h}^{\zeta^{t},t} &\gets \mathrm{exp}^{\zeta^{t}}_{\mathbf{o}}\left ( \mathrm{log}^{\zeta^{t-1}}_\mathbf{o}\left (\mathbf{h}^{\zeta^{t-1},t-1}\right )\right ), 
\end{aligned}
\end{equation}
where $\zeta^{t-1}$ and $\zeta^{t}$ are last epoch curvature and new curvature respectively, $\gamma$ is a weight parameter of estimated curvature, and $\mathbf{o}$ is the origin of tangent space $\mathcal{T}_{\mathbf{o}} \mathbb{H}^{n}$.

\textbf{ACE-Agent Reward Function $R^{t}_{\mathrm{ACE}}$}: 
We define the reward of ACE-Agent directly based on the performance on the specific task comparing with the last state: 
\begin{equation}
\label{eq:ACEreward}
R^{t}_{\mathrm{ACE}}=\mathcal{M}(\mathbf{h}^{\zeta^{t},t})-\mathcal{M}(\mathbf{h}^{\zeta^{t-1},t-1}),
\end{equation}
where $\mathbf{h}^{\zeta^{t},t}$ is the hyperbolic embedding vector with $\zeta$ and $\mathcal{M}(\cdot)$ is the evaluation metric of the downstream task. 
\\

\subsection{Hyperbolic Graph Neural Network Agent (HGNN-Agent)}
Hyperbolic Graph Neural Network Agent (HGNN-Agent) aims to learn node representations fusing graph structure and feature information in hyperbolic space with a given curvature. 
Similar to GNNs in Euclidean space, HGNNs also follow a message-passing scheme. 
Each layer $\ell$ of HGNNs transforms and aggregates the neighbours' hidden feature of previous layer  $\ell-1$  in the tangent space of the origin and then projects the result into a hyperbolic space with different curvatures: 
\begin{equation}\label{eq:HGNNLayer}
    \mathbf{h}^{\ell} = \sigma^{\otimes^{\zeta}} \left (\textsc{Agg}^{\zeta} \left ( \left ( \mathbf{W}^{\ell} \otimes^{\zeta^{\ell-1}} \mathbf{h}^{\ell-1}\right) \oplus^{\zeta^{\ell-1}} \mathbf{b}^{\ell}\right) \right),
\end{equation}
where $\oplus^{\zeta}$ and $\otimes^{\zeta}$ are the M\"obius vector addition and scalar-vector multiplication~\cite{HNN:GaneaBH18}, $\sigma^{\otimes^{\zeta}}$ is the hyperbolic non-linear activation function, and $\textsc{Agg}^{\zeta}$ is the hyperbolic neighborhood aggregation.  

Next, we introduce the four operations (i.e., feature mapping, linear transformation, neighbor aggregation, and activation) of an HGNN layer with variable curvature in detail. 

\textit{(1) Hyperbolic Feature Mapping with Variable Curvature. }
For feature mapping from Euclidean space to hyperbolic space, we can perform the exponential mapping with curvature $K=-1/\zeta^2$ to transform the node features according Eq.~\eqref{eq:exp} as: 
\begin{equation}\label{equ:featuremap}
    \mathbf{h}^{\zeta}_{\mathbb{H}}=\mathrm{exp}^{\zeta}_{\mathbf{o}}\left (\mathbf{x}\right )=\mathrm{exp}^{\zeta}\left  (\left  (0,\mathbf{x}\right )\right ) , 
\end{equation}
where $\mathbf{x} = (x_1, x_2, \dots, x_n) \in \mathbb{R}^{n}$ is a vector in Euclidean space, $(0,\mathbf{x})\!=\!(0, x_1, x_2, \dots, x_n)$ satisfies the Lorentzian scalar product $\langle (0,\mathbf{x}), 0 \rangle_{\mathcal{L}}=0$. The feature mapping from a hyperbolic space with curvature parameter $\zeta$ to another hyperbolic space with $\zeta^\prime$ is $\mathbf{h}^{\zeta^\prime} = \mathrm{exp}^{\zeta^\prime}_{\mathbf{o}}\left ( \mathrm{log}^{\zeta}_{\mathbf{o}}\left (\mathbf{h}^{\zeta}\right )\right )$. 


\textit{(2) Hyperbolic Linear Transformation. }
HGNNs leverage the exponential and logarithmic mapping to perform Euclidean transformations in the tangent space $\mathcal{T}_{\mathbf{o}} \mathbb{H}^{n}$. 
The hyperboloid matrix multiplication $\otimes^{\zeta}$ and addition $\oplus^{\zeta}$ are defined as: 
\begin{equation}\label{equ:lineartrans}
\begin{aligned}
      \mathbf{W} \otimes^{\zeta} \mathbf{h}^{\zeta}_{\mathbb{H}} &:= \mathrm{exp}^{\zeta}_{\mathbf{o}}\left (\mathbf{W} \cdot \mathrm{log}^{\zeta}_{\mathbf{o}}\left (\mathbf{h}^{\zeta}_{\mathbb{H}}\right )\right ),   \\
    \mathbf{h}^{\zeta}_{\mathbb{H}} \oplus^{\zeta} \mathbf{b} &:= \mathrm{exp}^{\zeta}_{\mathbf{h}^{\zeta}_{\mathbb{H}}}\left (P^{\zeta}_{\!\mathbf{o}\to \mathbf{h}^{\zeta}_{\mathbb{H}}}\left (\mathbf{b}\right )\right ),
\end{aligned}
\end{equation}
where $\mathbf{W}$ is the weight parameter, $\mathbf{b}$ is the bias parameter, and $P^{\zeta}(\cdot)$ is the parallel transformation from a uniform tangent space $\mathcal{T}_{\mathbf{o}} \mathbb{H}^{n}$ to the $\mathbf{h}^{\zeta}_{\mathbb{H}}$ tangent space $\mathcal{T}_{\mathbf{h}^{\zeta}_{\mathbb{H}}} \mathbb{H}^{n}$.

\textit{(3) Hyperbolic Neighborhood Aggregation. }
To facilitate the computation of weighted summation in hyperbolic space, most HGNNs leverage the logarithmic mapping to project and aggregate the node embeddings $\mathbf{h}^{\zeta}_{i \mathbb{H}}$ in the tangent space of each center point, and then map the average node embeddings back to the hyperboloid space by exponential mapping as introduced in Section~\ref{sec:background}. 
In this way, HGNNs can directly perform the Euclidean attention mechanism in tangent space, and the attention based aggregation in hyperbolic space can be defined as follows: 
\begin{equation}\label{eq:aggt}
\begin{aligned}
&w_{ij} = softmax_{j \in \mathcal{N}(i)}\left(\textsc{MLP}\left (\mathrm{log}^{\zeta}_{\mathbf{o}}\left (\mathbf{h}^{n,\zeta}_{i \mathbb{H}}\right )||\mathrm{log}^{\zeta}_{\mathbf{o}}\left (\mathbf{h}^{n,\zeta}_{j \mathbb{H}}\right )\right )\right ), \\
&\textsc{Agg}^{\zeta}\left (\mathbf{h}^{n,\zeta}_{i \mathbb{H}}\right ) = \mathrm{exp}^{\zeta}_{\mathbf{h}^{n,\zeta}_{i \mathbb{H}}}\left (\sum_{j \in \mathcal{N}(i)}w_{i j}\mathrm{log}^{\zeta}_{\mathbf{h}^{n,\zeta}_{i \mathbb{H}}}\left (\mathbf{h}^{n,\zeta}_{i \mathbb{H}}\right )\right ). 
\end{aligned}
\end{equation}


\textit{(4) Hyperbolic Activation with Variable Curvature. } 
Similar to the above operations, HGNNs also extend the Euclidean non-linear activation function to the hyperbolic space $\mathbb{H}^{n}$ by using the tangent space $\mathcal{T} \mathbb{H}^{n}$. 
We use the tangent space of the origin $\mathcal{T}_{\mathbf{o}} \mathbb{H}^{n}$ shared across hyperboloid manifolds with different curvatures. 
Hence the hyperbolic activation function $\sigma^{\otimes^{\zeta^\prime,\zeta}}$ with variable curvature is: 
\begin{equation}\label{equ:Hdist}
    \sigma^{\otimes^{\zeta^\prime,\zeta}}\left (\mathbf{h}^{n,\zeta}_{i \mathbb{H}}\right ) := \mathrm{exp}^{\zeta^\prime}_{\mathbf{o}}\left (\sigma\mathrm{log}^{\zeta}_{\mathbf{o}}\left (\mathbf{h}^{n,\zeta}_{i \mathbb{H}}\right )\right ). 
\end{equation}

\begin{algorithm}[!t]
\LinesNumbered
\label{alg}
\caption{ACE-HGNN.
} 
\KwIn{Graph $G$; 
Number of training epochs $E$; Initial curvature parameter $\zeta_0$ for each HGNN encoding layer; Exploration probability $\epsilon$.}
\KwOut{Predicted result of the downstream task.}
Initialize model parameters; \\
\For{$t = 1, 2, \cdots, E$}{
    Get state $S^{t}=(\zeta^{t-1}_{\mathrm{HGNN}}, \zeta^{t-1}_{\mathrm{ACE}})$;
    \\
    Take an action $A^{t}$ by the $\epsilon$-greedy policy;\\
    \tcp{HGNN-Agent}
    Calculate node embeddings $\mathbf{h}^{\zeta^{t},t}_{\mathrm{HGNN}}$ by Eq.~\eqref{eq:HGNNLayer};\\
    Get reward $R^t_{\mathrm{HGNN}}$ for HGNN-Agent by Eq.~\eqref{eq:HGNNreward};\\
    \tcp{ACE-Agent}
    Calculate node embeddings $\mathbf{h}^{\zeta^{t},t}_{\mathrm{ACE}}$ by Eq.~\eqref{eq:mapcurv};
    \\
    Get reward $R^t_{\mathrm{ACE}}$ for ACE-Agent by Eq.~\eqref{eq:ACEreward};\\
    \tcp{Update both agents}
    Calculate best policy for each agent
    $\pi_{\mathrm{HGNN}}^*(S')$ and $ \pi_{\mathrm{ACE}}^*(S')$ by Eq.~\eqref{eq:policy};
    \\
    Update HGNN-Agent and ACE-Agent via Nash Q-learning by Eq.~\eqref{eq:nashQ};\\
    
}
\end{algorithm}

Based on the above operations, we can build a multi-layer HGNN model as the architecture of HGNN-Agent. 
Then we formally define the state, action, and reward of the HGNN-Agent as follows.

\textbf{HGNN-Agent State $S^{t}_{\mathrm{HGNN}}$}: 
HGNN-Agent aims to learn the optimal node representations in the hyperbolic space with the given curvature. 
We define the HGNN-Agent state as:
\begin{equation}
\label{hgnnstate}
\begin{aligned}
    S^{t}_{\mathrm{HGNN}} = \left (\zeta^{t-1}_1,\zeta^{t-1}_2,\dots,\zeta^{t-1}_L\right ),
\end{aligned}
\end{equation}
where $\left (\zeta^{t-1}_1,\dots,\zeta^{t-1}_\ell\right )$ are the curvatures of $t-1$ epoch from ACE-Agent, and $L$ is the layer number of the HGNN. 

\textbf{HGNN-Agent Action $A^{t}_{\mathrm{HGNN}}$}: 
The action of HGNN-Agent is defined as whether to update the learned embeddings by taking the new curvatures. 
Formally, the action space of HGNN-Agent is defined as:
\begin{equation}
\label{hgnnaction}
A^{t}_{\mathrm{HGNN}}=\{\zeta^{t}_{\mathrm{HGNN}} \gets \zeta^{t}_{\mathrm{ACE}}, \ \zeta^{t}_{\mathrm{HGNN}} \gets \zeta^{t-1}_{\mathrm{HGNN}}\}.
\end{equation}

\textbf{HGNN-Agent Reward Function $R^{t}_{\mathrm{HGNN}}$}: 
Same as the ACE-Agent, the reward of HGNN-Agent is also defined based on the performance improvement on the specific task comparing with the last state: 
\begin{equation}
\label{eq:HGNNreward}
R^{t}_{\mathrm{HGNN}}=\mathcal{M}(\mathbf{h}^{t})-\mathcal{M}(\mathbf{h}^{t-1}),
\end{equation}
where $\mathcal{M}(\cdot)$ is the evaluation metric of the downstream task. 

\subsection{Multi-Agent Learning with Nash Q-Learning}
In this section, we introduce the collaborative learning of the HGNN-Agent and the ACE-Agent based on reinforcement learning. 
The objective is to converge the learning of the two agents to Nash equilibrium (i.e., all agents cannot independently update the learning result to improve the collaborative performance of the downstream task). 




We leverage Nash Q-learning~\cite{hu2003nash} to update the two agents and adopt an $\epsilon$-greedy policy with an exploration probability $\epsilon$ when taking actions. 
The ACE-Agent and HGNN-Agent share a global state $S$, and the Nash Q-learning optimization fits the Bellman optimality equation as below: 
\begin{equation}
\label{eq:nashQ}
\begin{aligned}
    NashQ_i\left (S'\right ) ~=~ & \pi_{\mathrm{HGNN}}^*\left (S'\right ) \pi_{\mathrm{ACE}}^*\left (S'\right ) Q_i\left (S'\right ),\\
    Q_i\left (S, A^t_{\mathrm{HGNN}}, A^t_{\mathrm{ACE}}\right ) \gets ~& Q_i\left (S, A^{t}_{\mathrm{HGNN}}, A^t_{\mathrm{ACE}}\right )\\
    & + \alpha\left (R^t_i + \beta NashQ_i\left (S'\right )\right. \\
    & \left.- Q_i\left (S, A^{t}_{\mathrm{HGNN}}, A^t_{\mathrm{ACE}}\right )\right ), 
\end{aligned}
\end{equation}
where $Q(\cdot)$ is the Q-function, $\alpha$ is learning rate, and $\beta$ is the discount factor. 

If the HGNN-Agent and the ACE-Agent have reached Nash equilibrium, the RL algorithm will stop, and the curvature parameter $\zeta$ will keep fixed during the next training process. 
\begin{equation}
\label{eq:policy}
\begin{aligned}
    \pi_{\mathrm{HGNN}}^*\left (S'\right ),\! \pi_{\mathrm{ACE}}^*\left (S'\right ) \gets Nash\left (Q_{\mathrm{HGNN}}\left (S'\right ),\! Q_{\mathrm{ACE}}\left (S'\right )\right ),
\end{aligned}
\end{equation}
where $\pi_{\mathrm{HGNN}}^*$ and $\pi_{\mathrm{ACE}}^*$ are the best policy of two agents, respectively. 
Intuitively, Eq.~\eqref{eq:policy} means that the RL algorithm finds the optimal curvature. The overall algorithm is given in Algorithm~\ref{alg}.

\section{Experiments}
\label{section 5}

\begin{table*}[t]
\caption{Summary of experimental results: “average score ± standard deviation” (\%).
\vspace{-0.5em}
}
\centering 
\resizebox{\textwidth}{!}{
\begin{tabular}{lccccccccccc}
\toprule
\textbf{Dataset} & \multicolumn{2}{c}{\textbf{Citeseer}} & \multicolumn{2}{c}{\textbf{Cora}} & \multicolumn{2}{c}{\textbf{Pubmed}} &
\multicolumn{2}{c}{\textbf{WebKB}} &
\multicolumn{2}{c}{\textbf{PPI}} &
\multirow{3}{*}{\textbf{Avg. Rank}}\\ 
\textbf{Hyperbolicity} $\delta$ & \multicolumn{2}{c}{$ \delta = 3.5 $} & \multicolumn{2}{c}{\textbf{$ \delta = 2.5 $}} & \multicolumn{2}{c}{\textbf{$ \delta = 2 $}} & \multicolumn{2}{c}{\textbf{$ \delta = 1 $}} &
\multicolumn{2}{c}{\textbf{$ \delta = 1 $}}\\ \cmidrule(r){2-11}
\textbf{Task} & LP & NC & LP & NC & LP & NC & LP & NC & LP & NC\\
\midrule
MLP                                     & 76.39±0.02 & 74.70±0.01 & 57.45±0.01 & 59.50±0.01 & 68.27±0.02 & 72.40±0.00 & 72.33±0.10 & 72.22±0.01 & 51.62±0.01 & 37.10±0.02 & 7.5 \\
HNN~\cite{HNN:GaneaBH18}                & 87.81±0.01 & 79.20±0.04 & 84.65±0.02 & 60.04±0.01 & 89.74±0.03 & 72.90±0.01 & 79.38±0.01 & 81.82±0.01 & 51.13±0.06 & 36.47±0.02 & 6.3 \\
\midrule
GCN~\cite{GCN}                          & 91.15±0.01 & \underline{81.94±0.00} & 90.42±0.03 & 81.93±0.01 & 75.17±0.03 & 76.50±0.00 & 88.60±0.02 & 70.71±0.00 & 82.03±0.05 & \underline{40.44±0.06} & 4.3 \\
GAT~\cite{GAT}                          & 91.21±0.10 & 80.41±0.03 & 92.44±0.14 & \underline{83.03±0.01} & 85.77±0.01 & \underline{78.30±0.02} & 83.77±0.08 & 63.64±0.02 & \underline{86.43±0.03} & 39.92±0.07 & 4.1 \\
GraphSAGE~\cite{hamilton2017inductive}  & 87.93±0.25 & 65.16±0.06 & 91.33±0.05 & 71.74±0.05 & 88.23±0.17 & 76.04±0.03 & 88.54±0.23 & 74.70±0.37 & 85.34±0.20 & 40.12±0.01 & 5.0 \\
\midrule
HGCN~\cite{HGCN_ChamiYRL19}             & \underline{93.90±0.02} & 77.60±0.02 & \underline{92.90±0.02} & 81.03±0.01 & 93.83±0.01 & 74.20±0.01 & 90.18±0.10 & 83.33±0.02 & 83.18±0.01 & 40.17±0.07 & 3.5 \\
$\kappa$GCN~\cite{bachmann2020constant} & 90.29±0.10 & 52.56±9.12 & 92.87±0.04 & 74.88±0.35 & \underline{93.88±0.08} & 73.52±0.08 & \underline{94.23±0.09} & \textbf{85.25±0.12} & 84.42±0.08 & 37.21±0.06 & 4.2 \\
\midrule
ACE-HGNN (Ours)   & \textbf{96.94±0.02} & \textbf{85.28±0.01} & \textbf{93.55±0.02} & \textbf{83.91±0.01} & \textbf{95.01±0.02} & \textbf{84.20±0.01} & \textbf{94.42±0.11} & \underline{83.38±0.01}  & \textbf{91.30±0.01} & \textbf{40.46±0.03} & \textbf{1.1} \\
\bottomrule
\end{tabular}}
\vspace{-1em}
\label{summary results}
\end{table*}

\begin{table}[!t]
\caption{Statistics of Datasets. }
\vspace{-0.5em}
\centering
\begin{tabular}{lrrrrr}
\toprule
\textbf{Dataset} & \textbf{\# Nodes} & \textbf{\# Edges} & \textbf{\# Features} & \textbf{\# Classes} & $\delta$\\
\midrule
Citeseer & 3,327   & 4,732   & 3,703 & 6   & 3.5 \\
Cora     & 2,708   & 5,429   & 1,433 & 7   & 2.5 \\
Pubmed   & 19,717  & 44,327  & 500   & 3   & 2 \\
WebKB    & 877     & 1,608   & 1,703 & 5   & 1 \\
PPI      & 14,755  & 228,431 & 50    & 117 & 1 \\
\bottomrule
\end{tabular}
\vspace{-1em}
\label{banchmark dataset}
\end{table}

In this section, we conduct comprehensive experiments to demonstrate the effectiveness and adaptability of ACE-HGNN on various datasets and tasks, and the results are shown in Table~\ref{summary results}. 
We further analyze the training efficiency, the embedding distortion, and the attention weights to investigate the expressiveness of ACE-HGNN. 

\subsection{Experimental Setup}
\textbf{Datasets.} 
We use a variety of open datasets including citation networks (Cora~\cite{sen2008cora}, Citeseer~\cite{GCN}, and Pubmed~\cite{namata2012PUBMED}), hypertext networks (WebKB~\cite{bachmann2020constant}) and protein-protein interaction networks (PPI~\cite{szklarczyk2016string}). 
The dataset statistics are shown in Table~\ref{banchmark dataset}. 
Since focusing on the hierarchy of the graph, we compute the Gromov’s $\delta$-hyperbolicity~\cite{jonckheere2008scaled,narayan2011large,adcock2013tree} for each graph dataset, where $\delta$ can measure how tree-like a graph is. 
A smaller $\delta$ indicates the graph is more hyperbolicity, and the graph becomes a tree when $\delta$ = 0. 

\textbf{Baselines.} 
We compare ACE-HGNN with a variety of baseline methods including neural network methods, Euclidean GNNs, and hyperbolic GNNs. 
For neural network methods, we consider Multi-Layer Perceptron (MLP) in Euclidean space and HNN~\cite{HNN:GaneaBH18} in hyperbolic space. 
For Euclidean GNNs, we consider GCN~\cite{GCN}, GAT~\cite{GAT}, and GraphSAGE~\cite{hamilton2017inductive}. 
For hyperbolic GNNs, we consider HGCN~\cite{HGCN_ChamiYRL19} and $\kappa$GCN~\cite{bachmann2020constant}. 

\textbf{Settings.} 
We set the common parameters as the number of layers $L$=2, the hidden layer and embedding dimension $dim$=16, $dropout$=0.5, the learning rate $lr$ varying from 1e-4 to 5e-4, and the weight decay is 5e-3. 
Espacially, for the hyperbolic models (HNN, HGCN, $\kappa$GCN, and our ACE-HGNN), the initial curvature $\zeta_0$=-1. 
The other parameters are set as the default values in their papers. 
For our ACE-HGNN, the $\epsilon$ in $\epsilon$-greedy policy is set to 0.9 initially, and it decays in each epoch until reaching $\epsilon_{min}$=0.1, the weight parameter of estimated curvature $\gamma$=0.2, the learning rate of Q-Learning $\alpha$ varying from 0.2 to 0.9, and the discount factor $\beta$=0.9. 
We follow the dataset split setting in~\cite{HGCN_ChamiYRL19}.

\subsection{Performance Comparison on Benchmark Datasets}
We evaluate our methods compared with baselines for two downstream tasks, link prediction (LP) and node classification (NC). 
Table~\ref{summary results} summarizes the performance of ACE-HGNN and the baseline methods on five datasets, where the best results are shown in bold and the second-best ones are underlined. 
We run five times and report the average results and standard deviations. 

\textit{For the link prediction task}, we sample a portion of nodes and edges from the network, marking them with identical numbers of positive tags and negative tags. 
We use Fermi-Dirac decoder~\cite{Krioukov2010Hyperbolic,NickelK17Poincare}, a generalization of sigmoid, to compute the connection probability between two nodes. 
We utilize ROC-AUC score to measure the performance. 
Compared with the second-best results, ACE-HGNN improves the ROC-AUC of link prediction by 3.04\% and 4.87\% on Citeseer and PPI, respectively. 
\textit{For the node classification task}, we directly classify nodes on the hyperboloid manifold using the hyperbolic multi-class logistic regression~\cite{HNN:GaneaBH18} and use the F1 score as the performance indicator. 
Compared with the second-best method, ACE-HGNN also improves the F1 score of node classification by 3.34\% and 5.90\% on Citeseer and Pubmed, respectively. 


\begin{figure}[!t]
	\centering
    \includegraphics[width=0.43\textwidth]{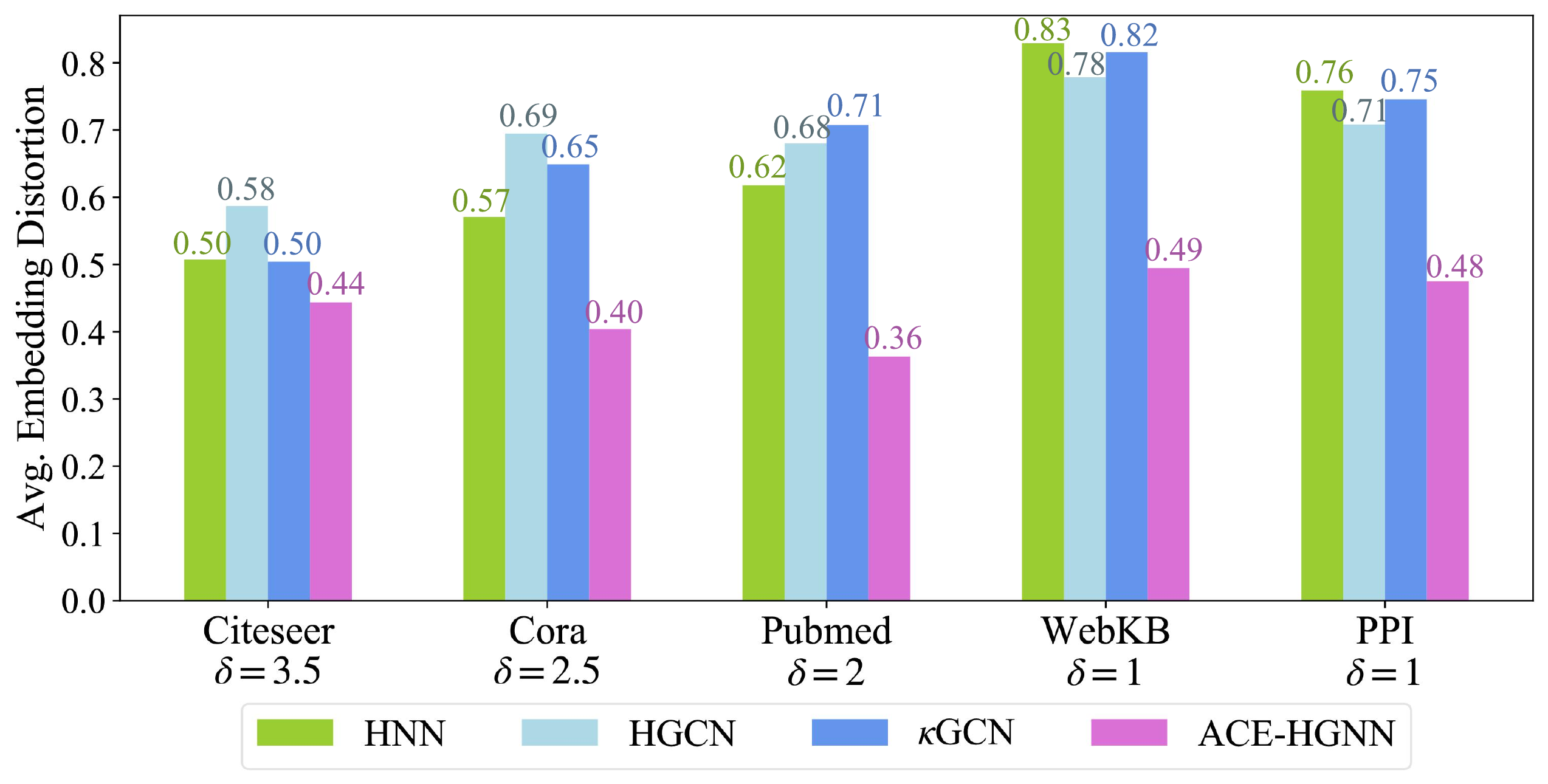}
    \vspace{-1em}
    \caption{Comparison of average embedding distortion.}
	\label{result:distortion}
    \vspace{-1.5em}
\end{figure}

Overall, we can get the following observations: 
(1) Our ACE-HGNN shows satisfying representative ability and achieves the best average performance among all datasets. 
(2) Generally, previous hyperbolic models (HNN, HGCN, and $\kappa$GCN) perform better on datasets with the higher hyperbolicity (lower $\delta$) but worse on datasets with lower hyperbolicity. 
The observations indicate that it is necessary to fuse hierarchical topology and feature information adaptively. 
Rather than taking the curvature as a hyper-parameter (as in $\kappa$GCN) or a learned parameter (as in HGCN), \textit{our ACE-HGNN quite benefits a lot from the adaptive curvature exploration mechanism}. 

\begin{figure*}[!t]
\centering
\subfigure[LP on Cora.]{
		\includegraphics[width=0.28\textwidth]{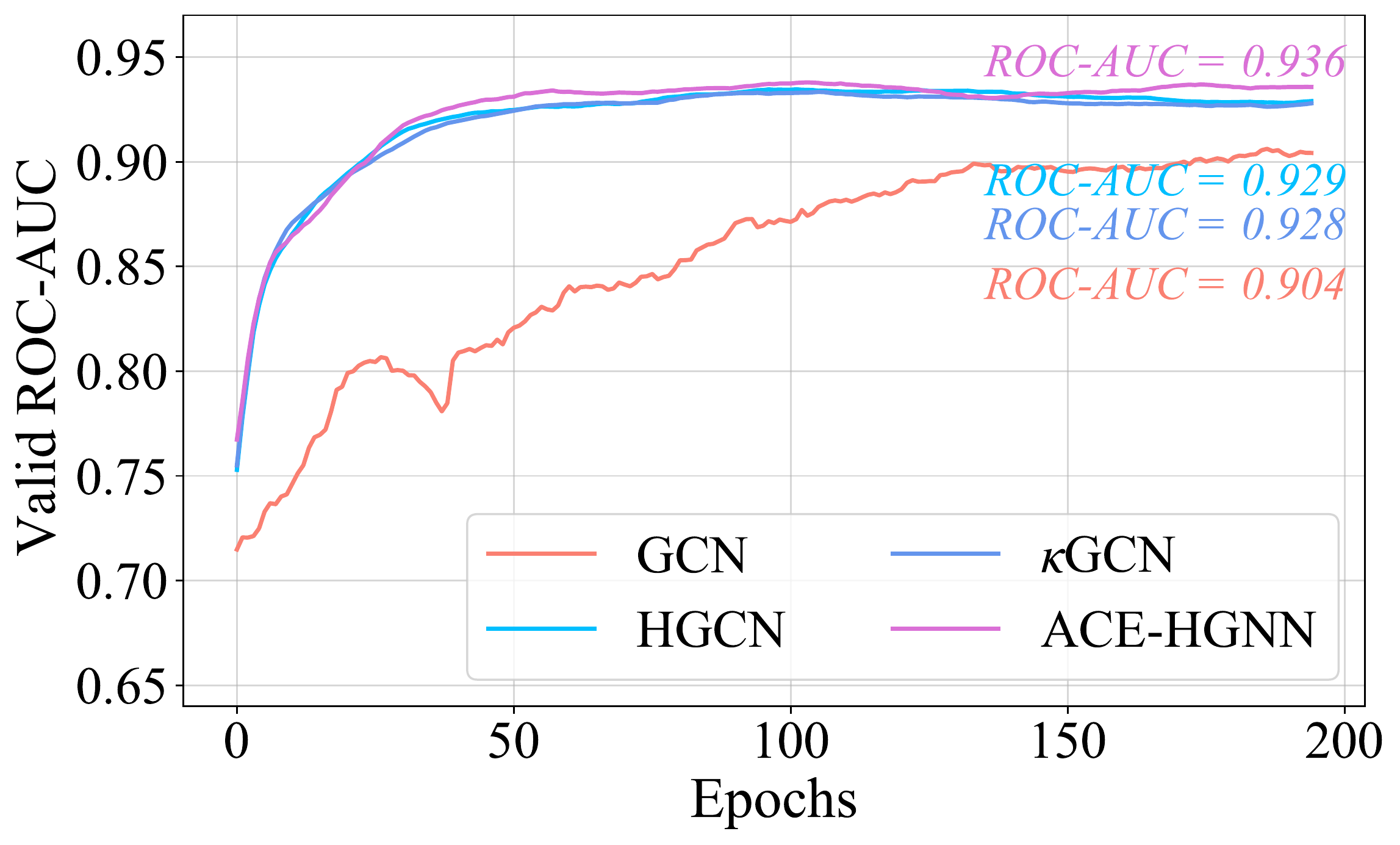}
	}
\hspace{0.5em}
\subfigure[LP on Pubmed.]{
		\includegraphics[width=0.28\textwidth]{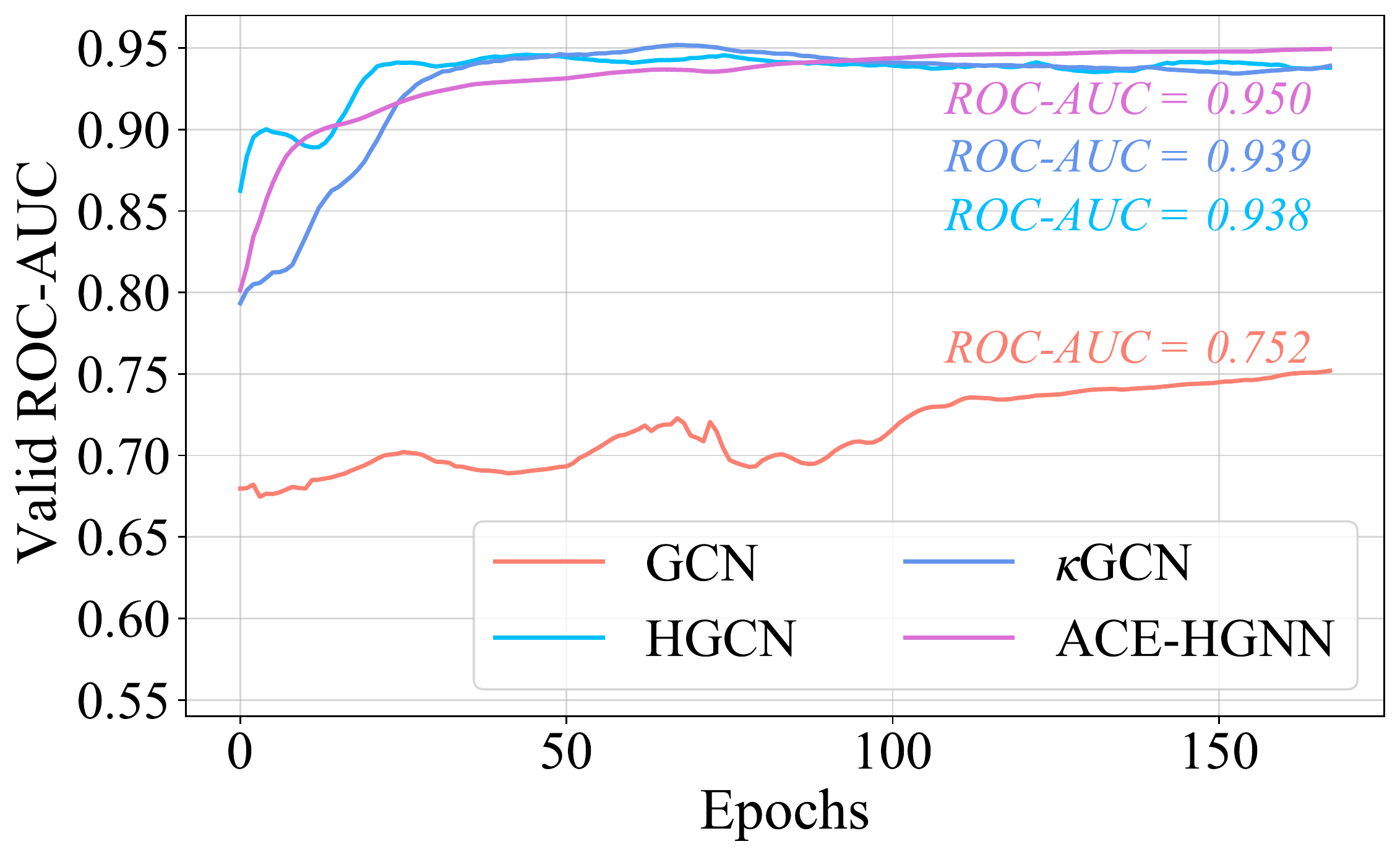}
	}
\hspace{0.5em}
\subfigure[LP on WebKB.]{
		\includegraphics[width=0.28\textwidth]{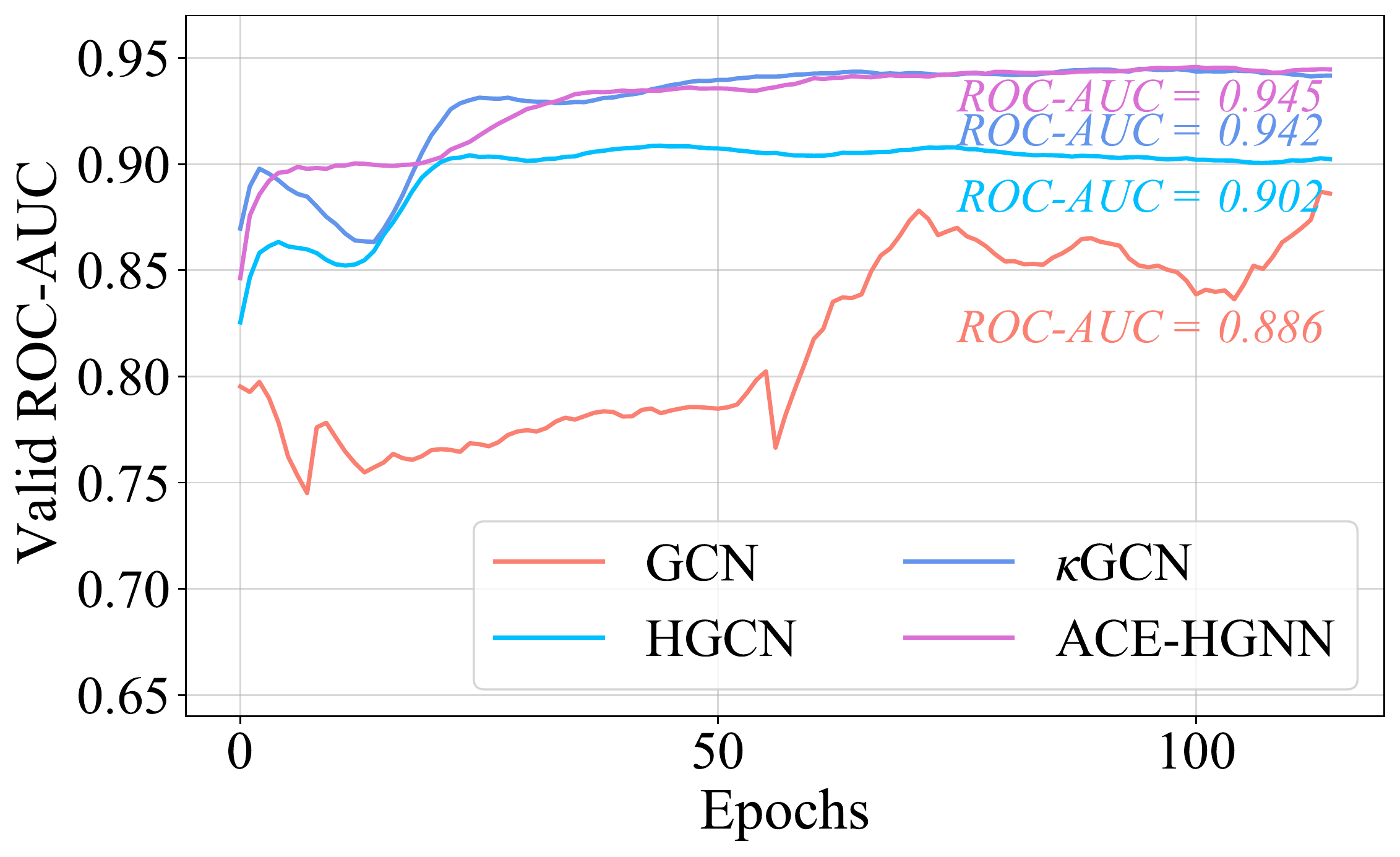}
	}
\vspace{-0.5em}
\caption{Learning curves of ROC-AUC scores of LP on validation set.}
\vspace{-1em}
\label{fig:loss}
\end{figure*}

\begin{figure*}[!t]
	\centering
	\subfigure[Curvature learning of LP on Cora.]{
		\includegraphics[width=0.28\textwidth]{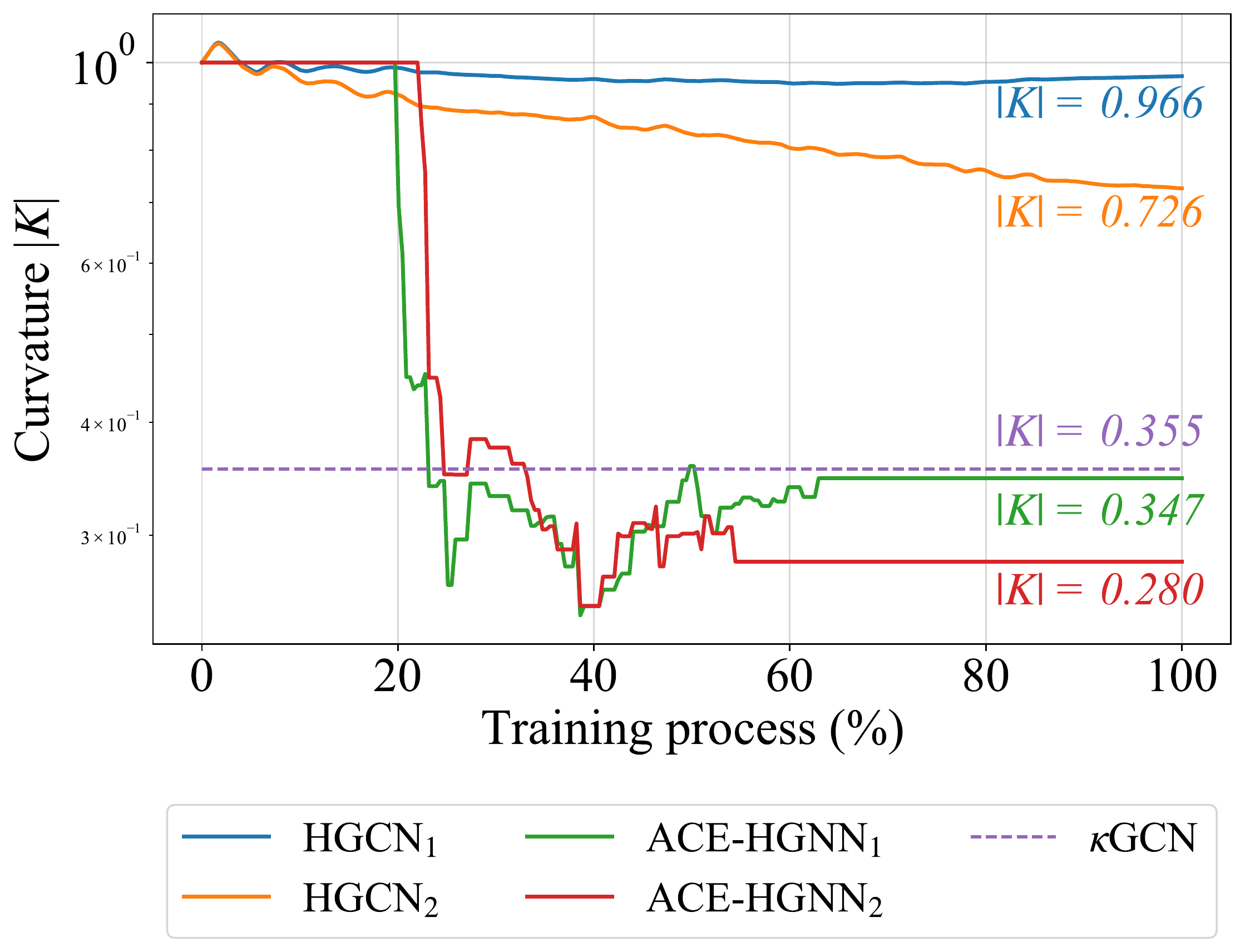}
	}
\hspace{0.5em}
	\subfigure[Curvature learning of LP on Pubmed.]{
		\includegraphics[width=0.28\textwidth]{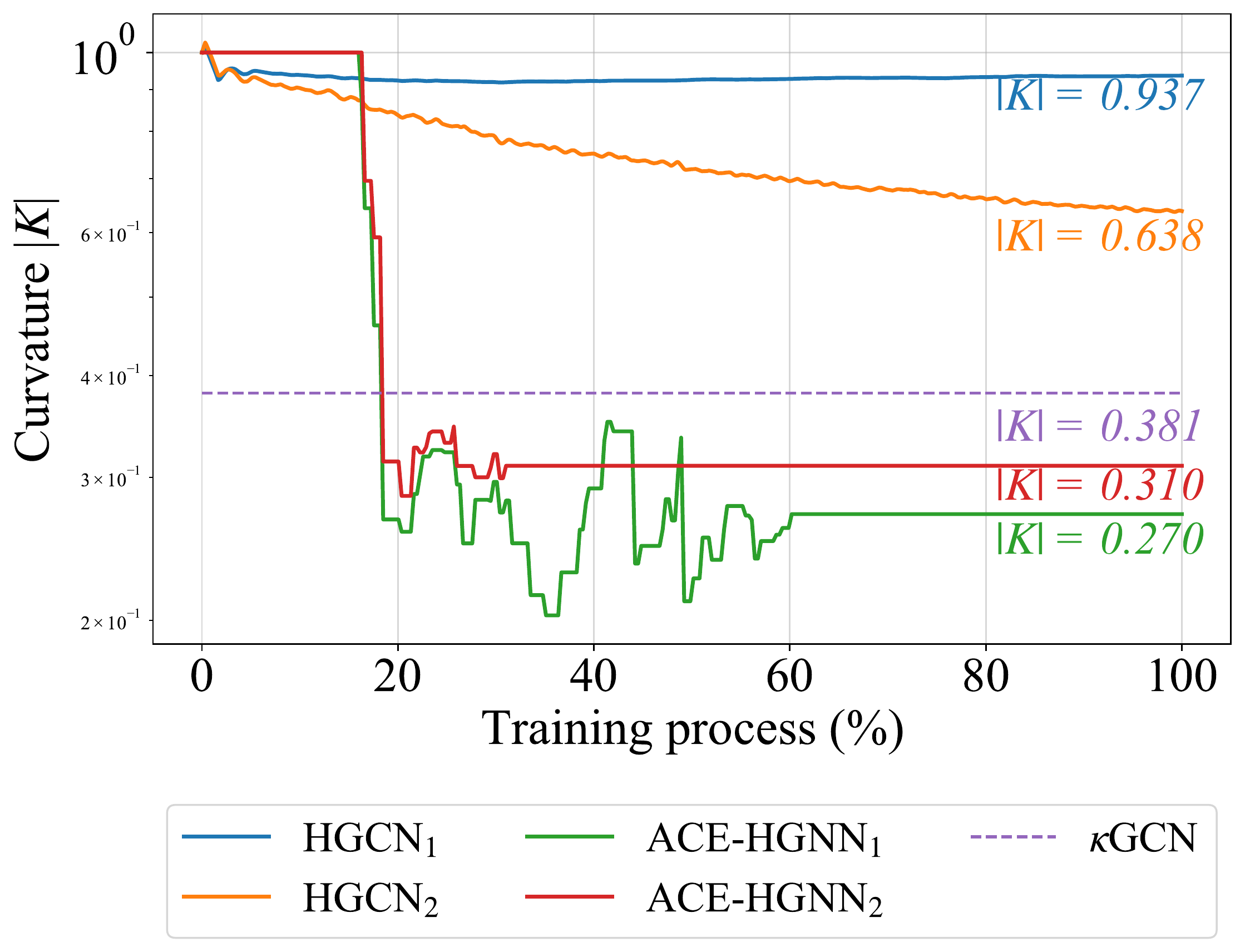}
	}
\hspace{0.5em}
	\subfigure[Curvature learning of LP on WebKB.]{
		\includegraphics[width=0.28\textwidth]{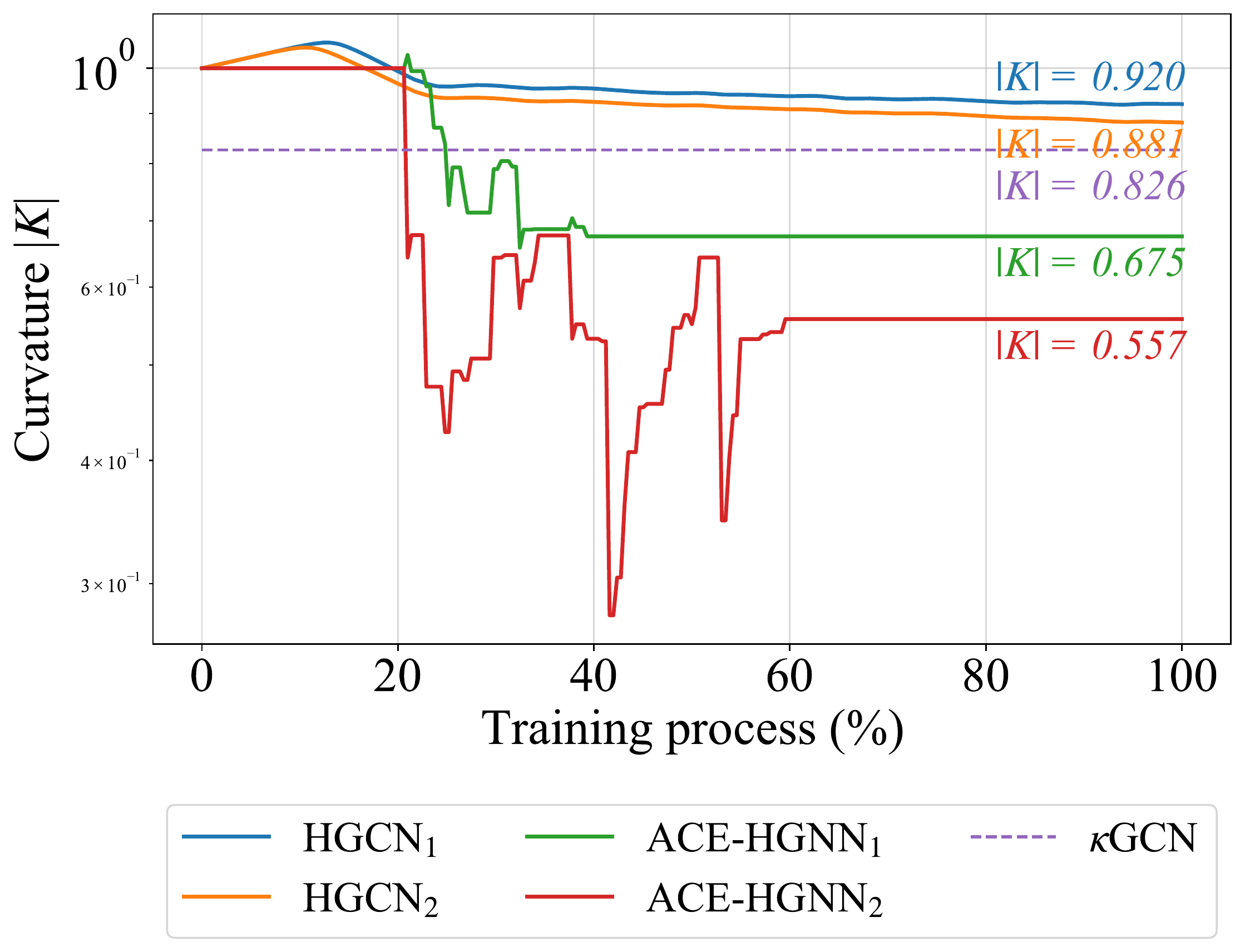}
	}
    \vspace{-0.5em}
	\caption{Curvature learning process of HGCN, $\kappa$GCN, and ACE-HGNN. }
    \vspace{-2em}
\label{fig:curvature}
\end{figure*}

\subsection{Further Analysis}
\textbf{Comparison of embedding distortion. }
In Section~\ref{sec:curv}, we discuss embedding distortion caused by hyperbolic graph learning models. 
Here we utilize the global average of embedding distortion (Definition~\ref{def:distortion}) to evaluate the distortions of benchmarks and our method on Cora and Pubmed. 
The average embedding distortions in link prediction task are shown in \figurename~\ref{result:distortion}, and a lower embedding distortion indicates better preservation of the graph structure. 
Obviously, ACE-HGNN has the lowest average embedding distortion among these models, indicating that the adaptive curvature exploration of our ACE-HGNN can effectively preserve the hierarchy of different graphs.

\textbf{Analysis of learning efficiency. }
We plot the training process of ACE-HGNN to verify the effectiveness of the proposed reinforcement learning framework in \figurename~\ref{fig:loss}, where each plot shows the changing trend of ROC-AUC value on valid set in the LP task during the whole learning process. 
On each dataset, ACE-HGNN converges within a few hundred epochs. 
It is obvious that the performance of ACE-HGNN quickly rises to the best. 
For example, on the Cora dataset, ACE-HGNN in 39 epochs achieves equivalent performance to HGCN in 100 epochs, giving us around 2.56x speedup.

\begin{figure*}[!t]
\centering
\subfigure[GCN]{
\includegraphics[width=0.18\linewidth]{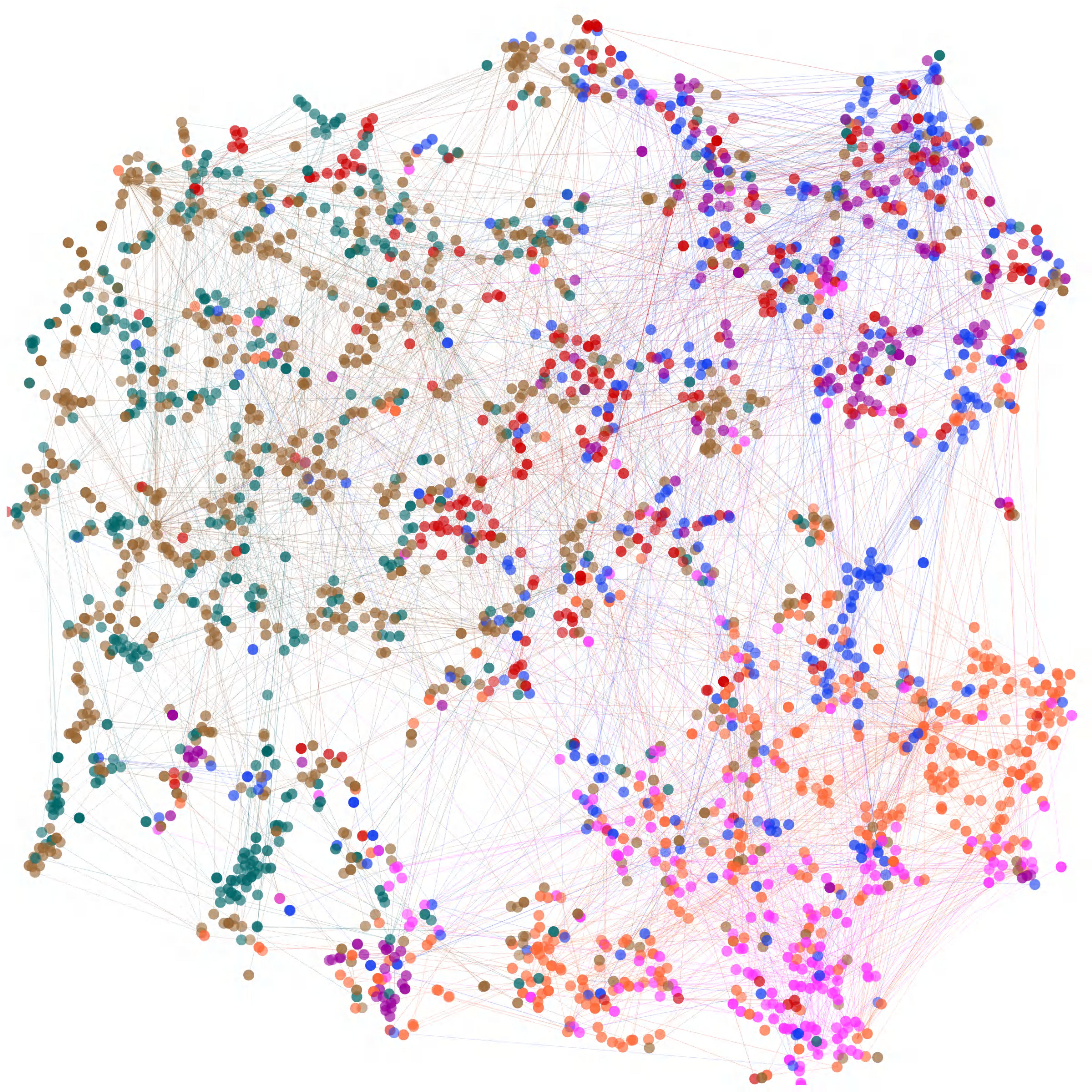}
}%
\hspace{2em}
\subfigure[GAT]{
\includegraphics[width=0.18\linewidth]{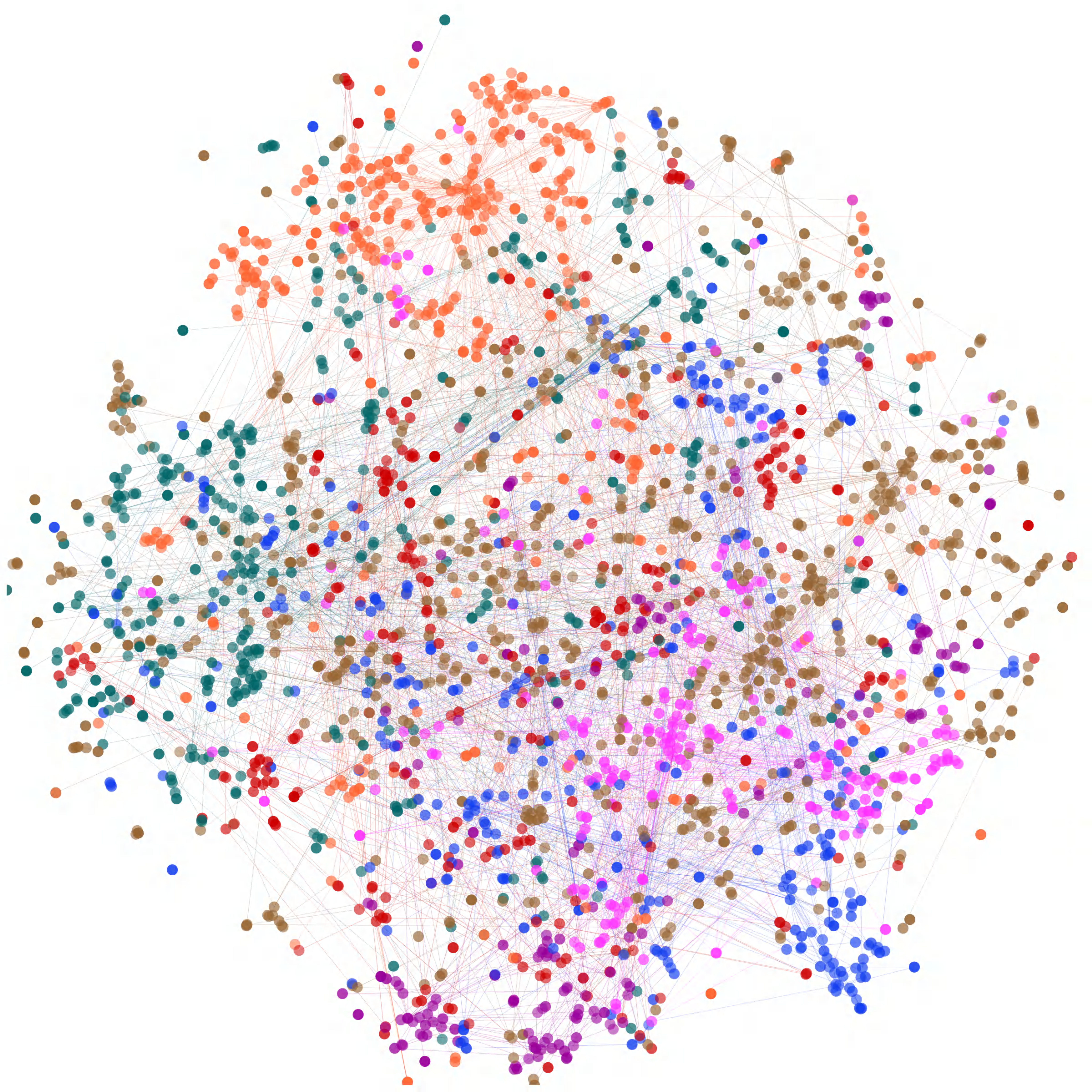}
}%
\hspace{2em}
\subfigure[HGCN ($K$ = -0.832)]{
\includegraphics[width=0.18\linewidth]{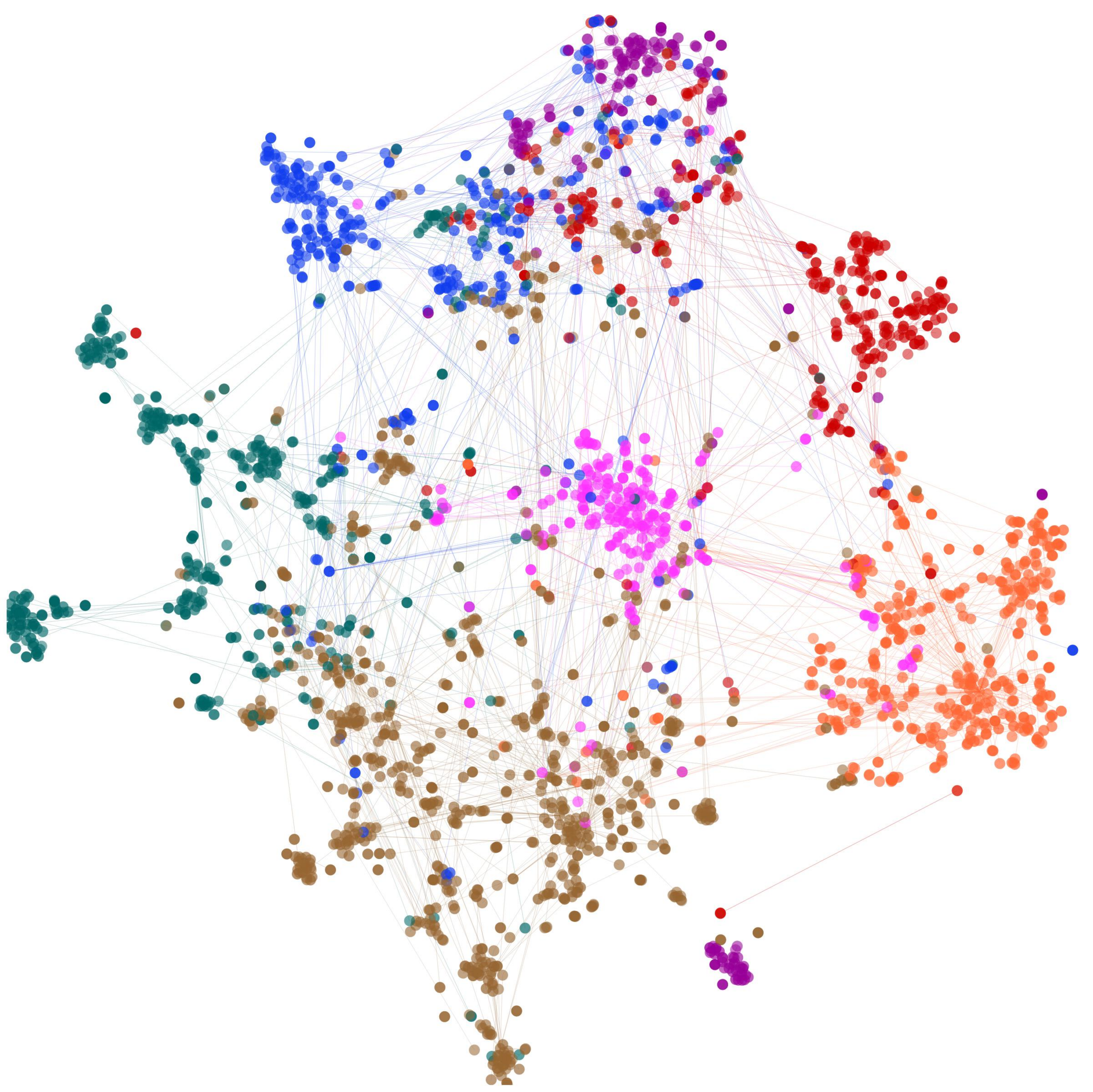}
}
\hspace{2em}
\subfigure[ACE-HGNN ($K$ = -0.436)]{
\includegraphics[width=0.18\linewidth]{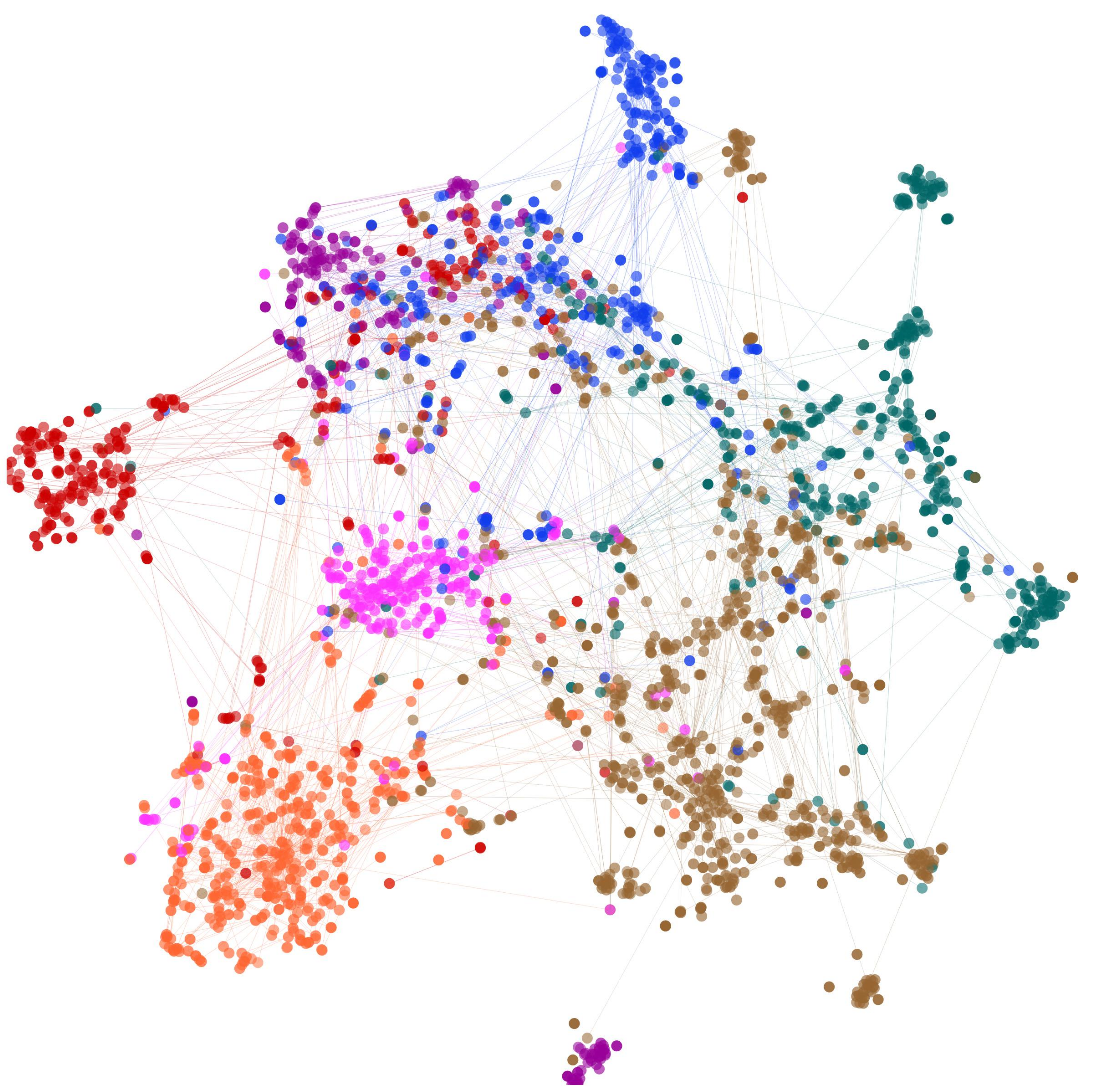}
}
\centering
\hspace{0.5em}
\caption{Visualization of node embeddings on Cora.}
\vspace{-1em}
\label{visualization1}
\end{figure*}

\begin{figure*}[!t]
\centering
\subfigure[Citeseer ($\delta$ = 3.5)]{
\includegraphics[width=0.14\linewidth]{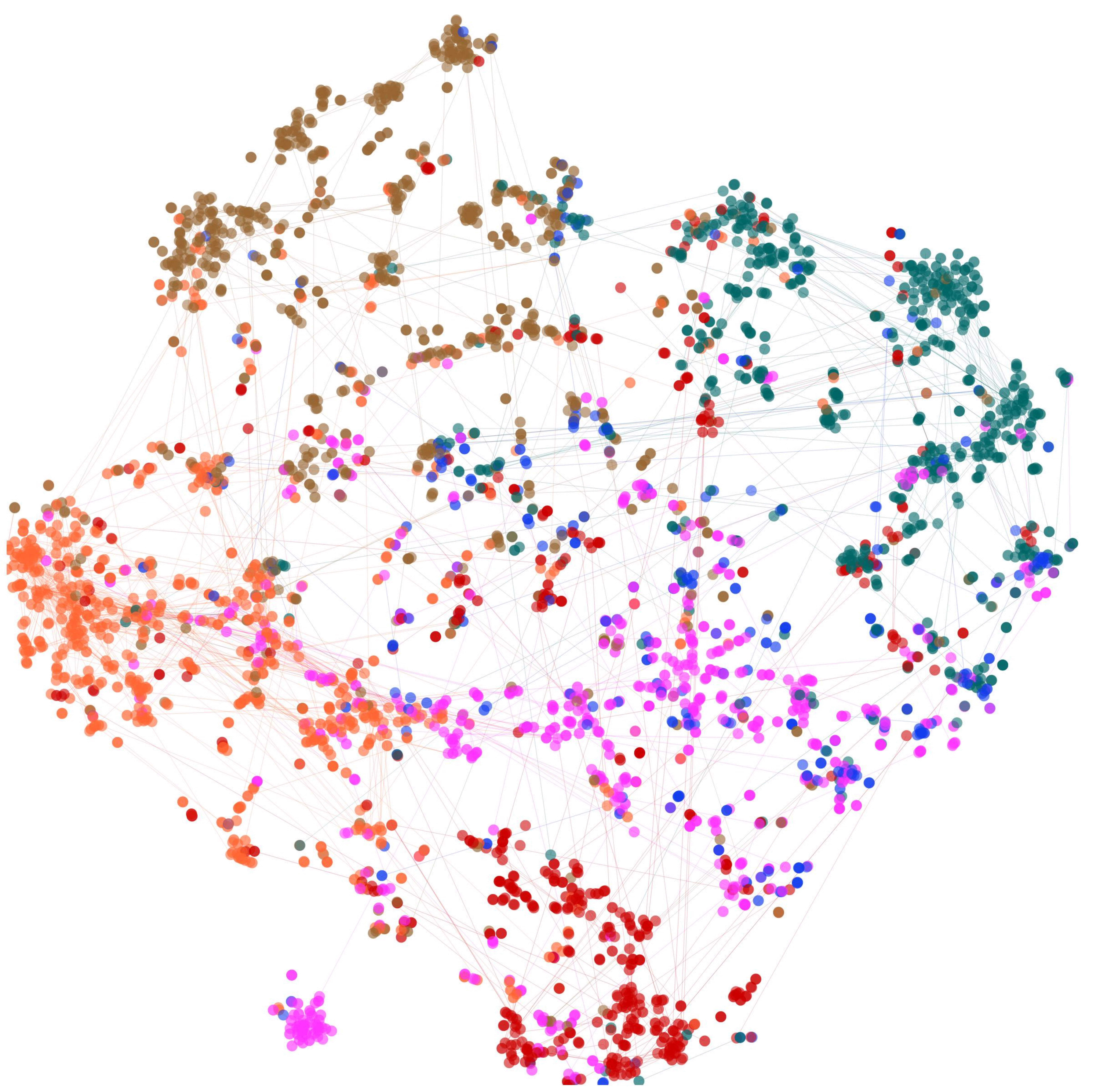}
\includegraphics[width=0.16\linewidth]{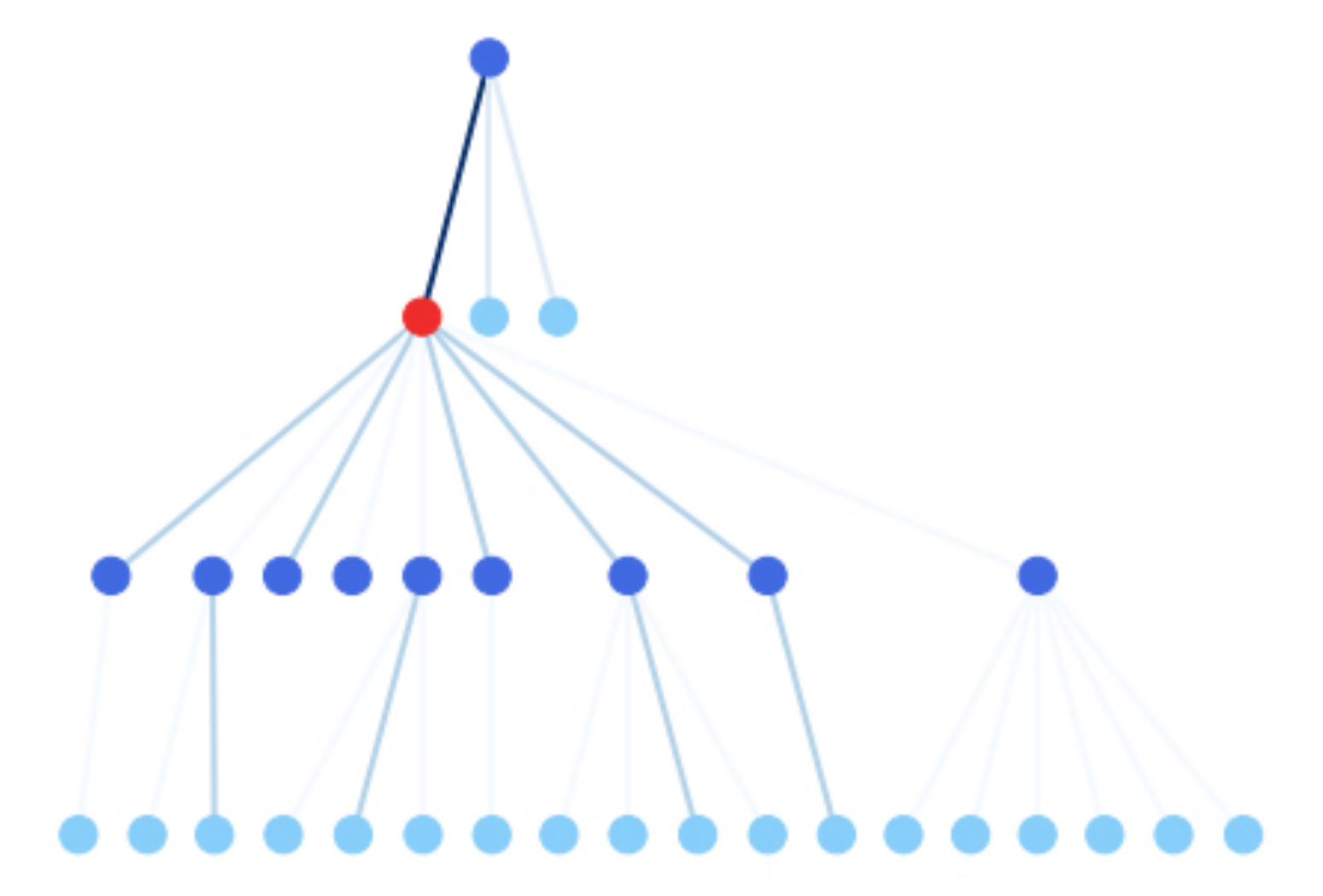}
}
\subfigure[Cora ($\delta$ = 2.5)]{
\includegraphics[width=0.14\linewidth]{ACE_cora_visualization.pdf}
\includegraphics[width=0.16\linewidth]{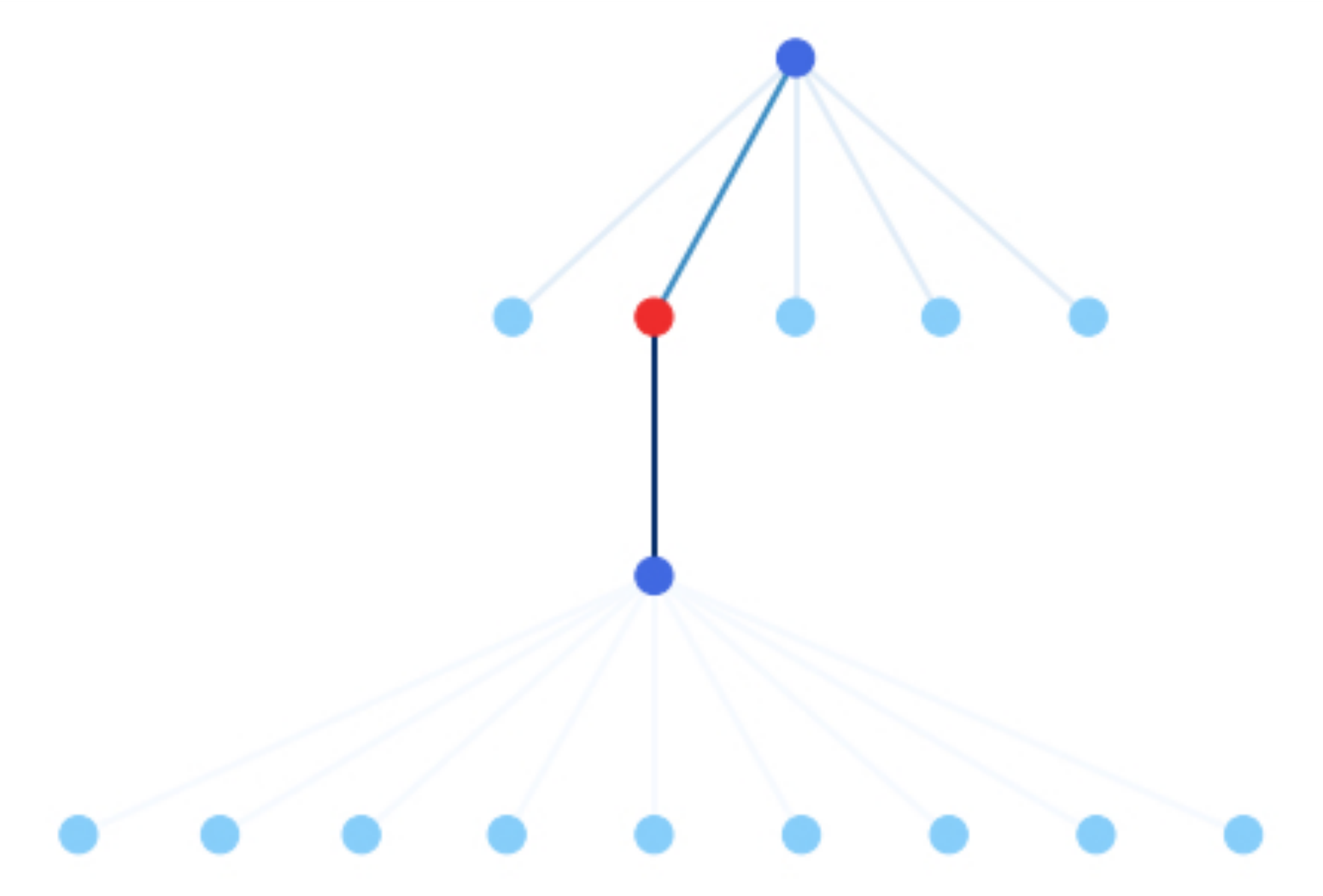}
}
\subfigure[WebKB ($\delta$ = 1)]{
\includegraphics[width=0.14\linewidth]{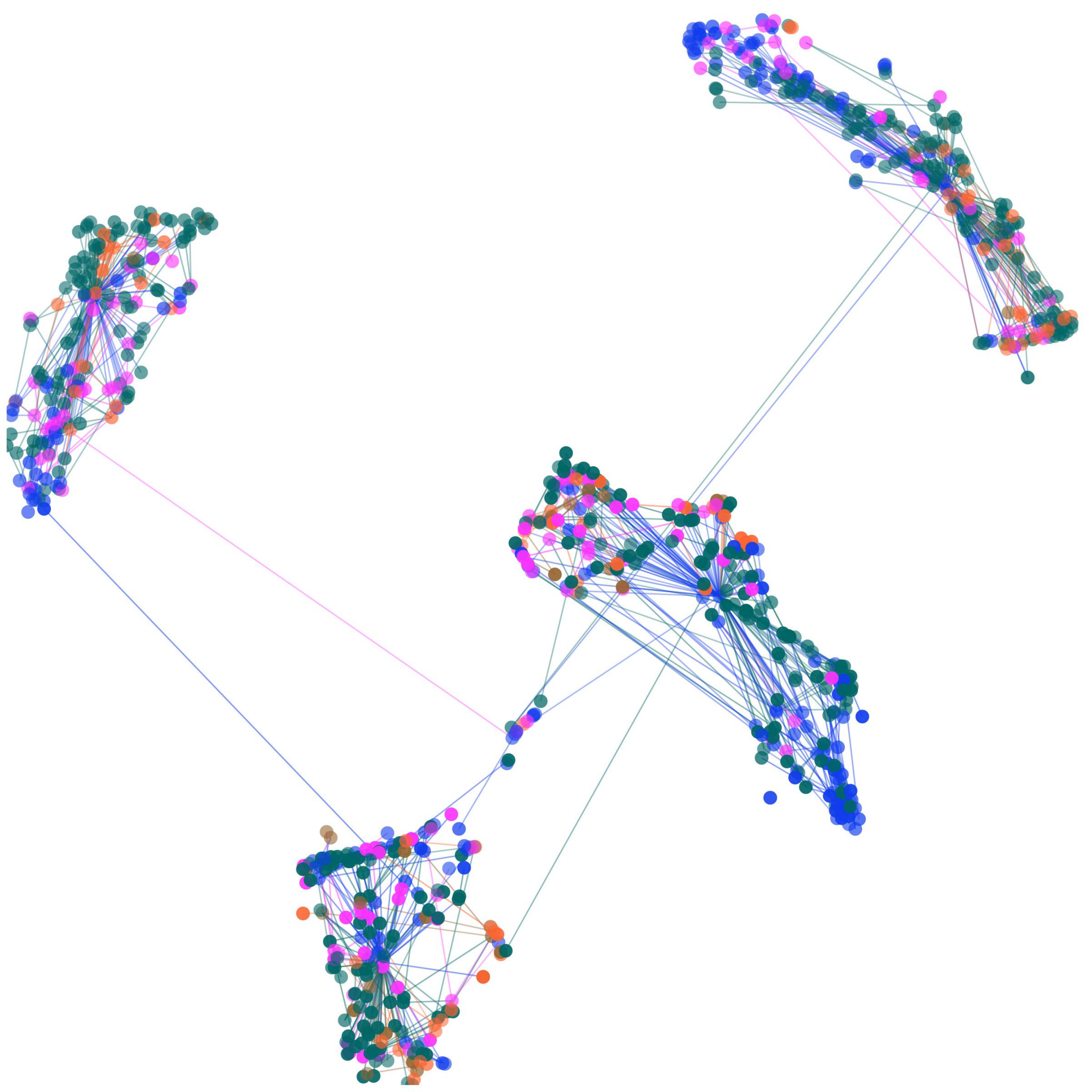}
\includegraphics[width=0.16\linewidth]{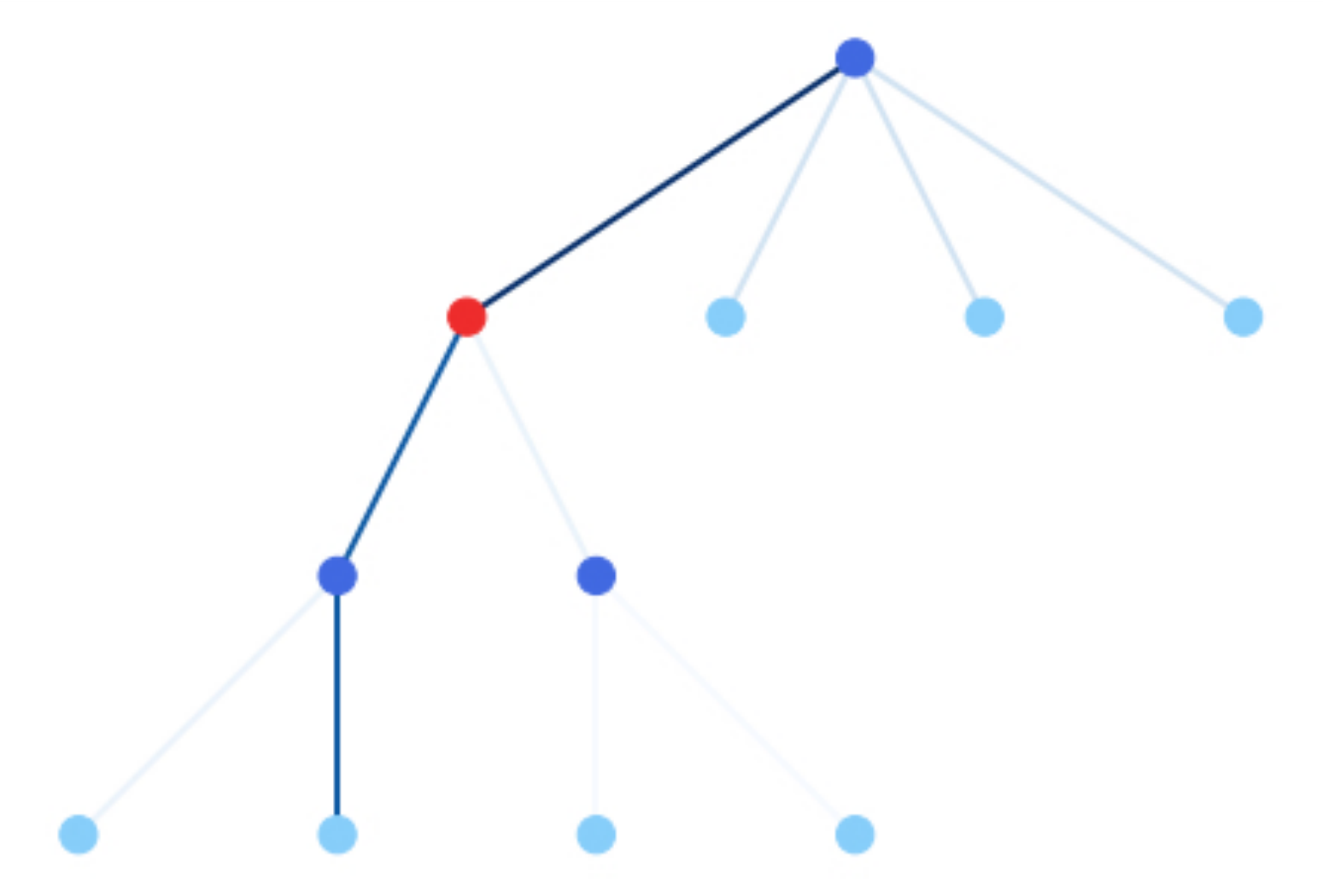}
}
\vspace{-0.5em}
\centering
\caption{Embeddings (left) and attention weights (right) visualization of ACE-HGNN on datasets with different $\delta$-hyperbolicity.}
\vspace{-1em}
\label{visualization2}
\end{figure*}

\textbf{Analysis of curvature learning.}
Existing HGNNs mainly learn the curvature in two ways: using trainable curvature as a model parameter (as in HGCN) or estimating an optimal curvature by the statistics (as in $\kappa$GCN).
The comparison results of the curvature learning process are shown in \figurename~\ref{fig:curvature}. 
Note that the curvature of $\kappa$GCN is pre-estimated and keeps fixed during the learning process (\textcolor{mypurple}{the purple dashed line}). 
Compared with HGCN, our RL-based ACE-HGNN can explore a larger scope of curvature in the learning process and converge to a better curvature. 
In addition, the curvature learned by ACE-HGNN finally converges near the estimated optimal curvature of $\kappa$GCN and is automatically fine-tuned to achieve better performance. 
We also observe that the curvatures of the two layers (ACE-HGNN$_1$ and ACE-HGNN$_2$) present a close relationship with competition and cooperation: (1) in several continuous epochs, the curvatures of two HGNN layers often choose opposite actions simultaneously or remain unchanged at the same time; (2) but on the scale of the entire training process, the overall trend of two layers is the same. 

\textbf{Embedding visualization. }
\figurename~\ref{visualization1} shows the embedding visualization of nodes in the test set of Cora using t-SNE~\cite{van2008visualizing}, where the color of node indicates its class. 
We can observe that HGCN and ACE-HGNN cluster nodes in a better way on separation compared with GCN and GAT. 
In contrast to HGCN, our ACE-HGNN indicates higher intra-class similarity and clearer distinct boundaries among different classes. 
This phenomenon indicates that the hyperbolic spaces with inappropriate curvature are "underfitting" or "overfitting" to the topology of the graph, which may reduce the performance in downstream tasks. 

\textbf{Analysis of attention weight. }
To further illustrate the effect of curvature on aggregating neighbor information, we visualize the attention weights of HGNN-Agent neighborhood aggregations on datasets with different $\delta$-hyperbolicity.  \figurename~\ref{visualization2} shows the corresponding embeddings and attention weights in the 2-hop neighborhood of an example node (red). 
The color darkness denotes the hierarchy of nodes and the intensity of edges between the neighbours denotes the attention weights. 
We observe that the center nodes are more concerned about their (grand)parents on datasets with lower $\delta$. 
In contrast to Citeseer, the aggregations on Cora and WebKB pay more attention to the high hierarchy of nodes. 

\section{Related Work}\label{sec:relatedwork}
\subsection{Graph Neural Networks and Reinforcement Learning}
Graph Neural Networks (GNNs)~\cite{GCN,GAT,hamilton2017inductive} are powerful deep representation learning methods for graphs and have been widely studied. 
Most of existing GNNs focus on preserving local topological information via neighbors or high-order proximity~\cite{GAT,hamilton2017inductive,li2021higher}, walks~\cite{Jin_Song_Shi_2020}, and motifs~\cite{monti2018motifnet}.
There are a few attempts to leverage RL to optimize the representation capability of GNNs on various scenarios~\cite{wang2019reinforcement,dou2020enhancing,hei2021hawk,peng2021reinforced,sun2021sugar,peng2021lime} .  
However, they learn representations in Euclidean space and cannot effectively capture the hierarchical topological information, which is of vital importance in real-world graphs.

\subsection{Hyperbolic Graph Representation Learning}
Hyperbolic geometric space was introduced into complex networks earlier to represent the small-world and scale-free of complex networks~\cite{Krioukov2010Hyperbolic,papadopoulos2012popularity}. 
With high capacity and hierarchical-structure-preserving ability, hyperbolic space is recently introduced into graph neural networks~\cite{HGNN_Qi,HGCN_ChamiYRL19,sun2021hyperbolic}. 
To the best of our knowledge, only a few studies~\cite{gu2019learning,bachmann2020constant} have discussed the correlation of the graph topological structure and curvature of non-Euclidean geometric embedding. 
These works are used to estimate the optimal fixed curvature through the statistics of the original graph data, and then applied to the learning model of hyperbolic graph representation. 

\section{Conclusion}
In this work, we propose ACE-HGNN, a novel adaptive curvature exploration hyperbolic graph neural network via multi-agent reinforcement learning. 
We are the first to introduce multi-agent reinforcement learning to hyperbolic graph representation learning. 
For graphs with different hierarchical structures and various downstream tasks, our multi-agent reinforcement learning based method can search an appropriate hyperbolic space with optimal curvature and learn good node representations simultaneously. 
Moreover, the adaptive curvature exploration can also intuitively give a reasonable interpretation of model learning, reflecting the model's preference for graph feature or structure information. 
We hope that our work will inspire the future development of hyperbolic geometric representation learning. 



\section*{Acknowledgment}
The authors of this paper were supported by the NSFC through grants (No.U20B2053, 61872022 and 62172443), State Key Laboratory of Software Development Environment (SKLSDE-2020ZX-12), and the ARC DECRA Project (No. DE200100964).
This work is also supported in part by NSF under grants III-1763325, III-1909323,  III-2106758, and SaTC-1930941.




\bibliographystyle{IEEEtran}
\bibliography{reference}
%



\end{document}